\documentclass[twoside,11pt]{article}
\usepackage{jmlr2e}

\usepackage{morefloats}

\usepackage{graphicx}
\usepackage{epstopdf}
\usepackage{float}
\usepackage{latexsym}
\usepackage{multicol}
\usepackage{algorithmic}
\usepackage{pbox}
\usepackage{multirow}
\usepackage{hhline}
\usepackage{color}

\usepackage{amssymb}
\usepackage{amsmath}
\usepackage[ruled]{algorithm2e}

\makeatletter
\newcommand*{\indep}{%
  \mathbin{%
    \mathpalette{\@indep}{}%
  }%
}
\newcommand*{\nindep}{%
  \mathbin{
    \mathpalette{\@indep}{\not}
  }%
}
\newcommand*{\@indep}[2]{%
  \sbox0{$#1\perp\m@th$}
  \sbox2{$#1=$}
  \sbox4{$#1\vcenter{}$}
  \rlap{\copy0}
  \dimen@=\dimexpr\ht2-\ht4-.2pt\relax
  \kern\dimen@
  {#2}%
  \kern\dimen@
  \copy0 
}

\makeatother

\begin{document}

\title{Causality-based Feature Selection:  Methods and Evaluations}
\author{\name Kui Yu \email yukui@hfut.edu.cn \\
         \name Xianjie Guo \email Xianjie\_reasonless@163.com\\
\addr School of Computer Science and Information Engineering\\
      Hefei University of Technology, Hefei, 230601, China\\
         \name Lin Liu \email Lin.Liu@unisa.edu.au \\
          \name Jiuyong Li \email Jiuyong.Li@unisa.edu.au \\
\addr School of Information Technology and Mathematical Sciences\\
      University of South Australia, Adelaide, 5095, SA, Australia\\            
 \name Hao Wang \email jsjxwangh@hfut.edu.cn \\
          \name Zhaolong Ling \email z\_dragonl@163.com \\
\addr School of Computer Science and Information Engineering\\
      Hefei University of Technology, Hefei, 230601, China\\
\name Xindong Wu  \email wuxindong@mininglamp.com\\
\addr  Mininglamp Academy of Sciences, Mininglamp Technology, Bejing, 100084, China
}




\editor{}
\maketitle

\begin{abstract}

Feature selection is a crucial preprocessing step in data analytics and machine learning.  Classical feature selection algorithms select features based on the correlations between predictive features and the class variable and do not attempt to capture causal relationships between them. It has been shown that the knowledge about the causal relationships between features and the class variable has potential benefits for building interpretable and robust prediction models, since causal relationships imply the underlying mechanism of a system.
Consequently, causality-based feature selection has gradually attracted greater attentions and many algorithms have been proposed.
In this paper,  we present a comprehensive review of recent advances in causality-based feature selection.
To facilitate the development of new algorithms in the research area and make it easy for the comparisons between new methods and existing ones, we  develop the first open-source package, called CausalFS, which consists of most of the representative causality-based feature selection algorithms (available at https://github.com/kuiy/CausalFS). Using CausalFS, we conduct extensive experiments to compare the representative algorithms with both synthetic and real-world data sets. Finally, we discuss some challenging problems to be tackled in future causality-based feature selection research.
\end{abstract}


\section{Introduction}\label{sec1}

Feature selection plays an essential role in high-dimensional data analytics ~\citep{guyon2003an,aliferis2010local1,li2017feature,brown2012conditional,wu2013online} and it is widely employed in all kinds of machine learning solutions. Feature selection is to find a subset of features
from a large number of predictive features for building predictive models for a target or class variable of interest. For example, gene (i.e., feature) selection can identify a small number of informative genes from a high-dimensional gene data set for predicting a disease or directing experimental studies to validate the identified genes (as genetic factors of a disease) in laboratories. Now feature selection is more critical than ever, since a data set with high-dimensionality has become ubiquitous in various applications~\citep{zhai2014the,li2017challenges,yu2016scalable}. In the previous example,  a gene expression data set may easily have more than 10,000 predictive features~\citep{saeys2007a}. For another example, the Web Spam Corpus 2011 collected approximately 16 million predictive features for malicious web detection~\citep{wang2012evolutionary}.  Almost all machine learning methods may not directly work on data sets of such high dimensionality without feature selection.
As a result, in the last two decades, feature selection has been well studied and has achieved great success in reducing computational costs of learning and improving the generalization ability of predictive models~\citep{li2017feature}.

Existing feature selection methods can be broadly categorized into  filter, wrapper, and embedded methods. A filter method is independent of a predictive model, whereas the other two types of methods are predictive model dependent. Due to their independence of predictive models, filter methods are able to achieve fast processing speed and have no bias on specific  predictive models.  With the rapid increase of high dimensional data, filter methods have been attracting more attentions than ever.  In this paper,  we focus on causality-based feature selection, an emerging successful type of filter methods.
In feature selection, a feature is considered as a strongly relevant feature, or a weakly relevant feature, or an irrelevant feature with respect to a class variable of interest~\citep{kohavi1997wrappers}. A classical feature selection method aims to find a subset of relevant features based on the correlations between (predictive) features and the class variable~\citep{guyon2003an}.
In general, correlations do not capture the causal relationships between features and the class variable, but only their co-occurrences.
Recent studies have shown that causal features may provide the following potential benefits in feature selection for classification~\citep{guyon2007causal,aliferis2010local1}.

\begin{itemize}

\item  Causal features can improve the interpretability of predictive models~\citep{hofman2017prediction,ribeiro2016why}. Correlations capture only the co-occurrence of features and the class variable, hence the selected features often do not provide a convincing interpretation for predictions. For example,  a strong correlation between \emph{yellow fingers} (of a smoker) and   \emph{lung cancer} may be found in patient records, making \emph{yellow fingers} a good predictive feature of  \emph{lung cancer}. However, clearly \emph{yellow fingers} is not a reasonable interpretation at all for  \emph{lung cancer}. In fact, the causes of  \emph{lung cancer}, such as  \emph{smoking}, can provide a reasonable interpretation for the prediction of lung cancer.

\item  Causal features can improve the robustness of  predictive models~\citep{athey2017beyond,peters2016causal,zhang2015multi,magliacane2018domain}. Causal relationships imply the underlying mechanism about the class variable 
 and thus they are persistent across different settings or environments. For example, we want to build a predictive model to diagnose  \emph{lung cancer} using historical patient data. Based on the historical data, a predictive model built using non-causal features such as yellow fingers may not produce good predictions. This is because nowdays smokers are very careful to hide their smoking habit and do not leave yellow stain on their fingers, and hence the distribution of symptoms presented by patients may become different from that of the historical data.  In contrast, if the causes of  \emph{lung cancer} of patients (such as  \emph{smoking})  were selected as the predictive features, a model built on the historical data will be robust.

\end{itemize}

\begin{figure}
\centering
\includegraphics[height=1.6in,width=3.2in]{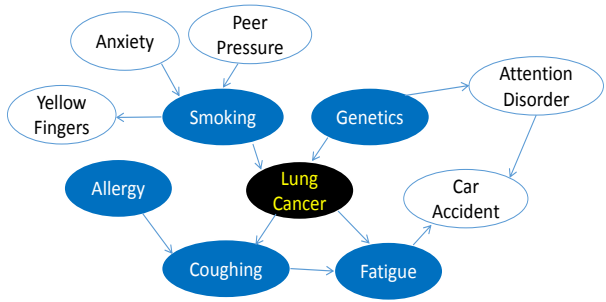}
\caption{An example of a MB in a lung-cancer Bayesian network}
\label{fig2-1}
\end{figure}

In recent years,  causality-based feature selection has gradually attracted more and more attentions from both machine learning and causal discovery domains~\citep{aliferis2010local1,guyon2007causal,borboudakis2019forward,yu2018unified}.
Causality-based feature selection methods  identify a Markov boundary (MB) or a subset of the MB such as parents and children (PC) (i.e., direct causes and direct effects) of the class variable from a data set.
The notion of MB was proposed in the context of a Bayesian network (BN)~\citep{pearl2014probabilistic}.
If the model generating the data set can be faithfully represented by a BN, then the MB of the class variable is unique and consists of the parents,  children, and spouses (i.e., other parents of the class variable's children) of the class variable.
Figure~\ref{fig2-1} gives an example of an MB in the BN of lung cancer~\citep{guyon2007causal}. The MB of \emph{Lung cancer} includes  \emph{Smoking} and \emph{Gentics} (parents), \emph{Coughing} and \emph{Fatigue} (children), and \emph{Allergy} (spouse).
As can be seen in Figure~\ref{fig2-1},  the MB of a class variable implies the local causal relationships between the class variable and the features (variables) in its MB.
Most importantly, under certain assumptions (to be discussed in Section~\ref{sec2}), the MB of the class variable is the minimal feature subset with maximum predictivity for classification since all other features are probabilistically independent of the class variable conditioning on its MB~\citep{pearl2014probabilistic,koller1996toward,tsamardinos2003towards}.
 

Since causality-based feature selection explicitly induces the local causal relationships around the class variable and has theoretical guarantees,
it becomes a promising direction~\citep{guyon2007causal,aliferis2010local1}.
In the past decade, many efforts have been made on causality-based feature selection and many algorithms have been proposed (without learning an entire BN structure involving all features in a data set)~\citep{aliferis2010local1,borboudakis2019forward,yu2018unified}.
The developed causality-based feature selection algorithms provide a new and complementary algorithmic methodology to enrich feature selection, especially for achieving explainable and robust machine learning methods.

To advance the research in causality-based feature selection, a comprehensive review of the state-of-the-art techniques in this area is in need. However, so far, there has not been such a review available.
Aliferis et al.~\citep{guyon2007causal,aliferis2010local1,aliferis2010local2} proposed a general local learning framework for causality-based feature selection, which are focused on three specific causality-based feature selection algorithms (e.g., MMMB, HITON-MB, and semi-HITON-MB) and their extensions to the BN structure learning, but the work is not a survey paper. 
Guyon et al.~\citep{guyon2007causal}  presented a comparison of the motivations and pros/cons of causality-based and classical feature selection approaches at the conceptual level, but again they did not provide a survey of causality-based feature selection algorithms.  Recently, Yu et al.~\citep{yu2018unified} discussed some representative causality-based and classical feature selection algorithms, but the work was focused on developing a unified view to link causality-based feature selection with classical feature selection instead of presenting an extensive review of causality-based feature selection algorithms. There have been some recent reviews on causal inference, such as~\citep{guo2018a,glymour2019review,zhang2017learning}, but they mainly focused on the advances on learning causal relations between features.
Meanwhile almost all reviews regarding feature selection focused on classical feature selection methods in the past decades~\citep{guyon2003an,brown2012conditional, li2017challenges,li2017feature}. 
In summary, so far there is little work on a comprehensive review of causality-based feature selection algorithms.

In addition,  there is no any open-source toolbox/package which implements existing causality-based feature selection algorithms. An open-source toolbox plays a crucial role for facilitating the development of new algorithms and making comparisons between the new methods and existing ones easy, and it may further promote both scientific and practical studies in machine learning and causal discovery. 

In this paper, we make the following contributions to fill the gaps discussed above.

\begin{itemize}

\item We extensively review the causality-based feature selection methods, including the two types of causality-based feature selection methods, constraint-based and score-based methods, and the algorithms of distinguishing causes and effects. To the best of our knowledge, this is the first attempt on presenting an extensive survey of causality-based feature selection and its recent advances.


\item We develop the first comprehensive open-source package that implements the representative and state-of-the-art causality-based feature selection algorithms. The package is written in C language, easy to use, and completely open source.

\item  We conduct a comprehensive empirical evaluation on representative causality-based feature selection algorithms with both synthetic and real-world data sets. 


\end{itemize}

The rest of the paper is organized as follows. Section~\ref{sec2} gives basic background knowledge. Section~\ref{sec3}  reviews  constraint-based methods. Section~\ref{sec4} reviews  score-based methods.
Section~\ref{sec5}  discusses the algorithms for distinguishing causes from effects using the outputs of causality-based feature selection algorithms.
Section~\ref{sec6} presents the open-source package of causality-based feature selection. Section~\ref{sec7} reports the  evaluation results.  Section~\ref{sec8}  concludes the paper and discusses some open problems.

\section{Markov boundary and Causality-based feature selection}\label{sec2}

In this section, we first briefly introduce the background knowledge of MB and BN, then we link MB with feature selection.

\subsection{ Bayesian network, Markov boundary, and causality-based feature selection}

Let $C$ be a class variable and $F=\{F_1,F_2,\cdots,F_M\}$ be a feature set including $M$ distinct features.
We use $F_i\indep F_j|S$, where $i\neq j$ and $S\subseteq F\setminus\{F_i, F_j\}$,  to denote that $F_i$ is conditionally independent of $F_j$ given feature set  $S$, and $F_i\nindep F_j|S$  to represent that $F_i$ is conditionally dependent on $F_j$ given $S$.

Let $S$ be any set of variables within $V$, we use $S\setminus V_i$ as the shorthand of  $S\setminus \{V_i\}$ and $S\cup V_i$ as the shorthand of  $S\cup \{V_i\}$.
Let $V=F\cup C=\{V_1, V_2, \cdots, V_{M+1}\}$, $V_i=F_i\ (1\leq i\leq M)$, and $V_{M+1}=C$. Let $P(V)$ be the joint probability distribution over $V$ and $G=(V, E)$ represent a directed acyclic graph (DAG) with nodes $V$ and edges $E$, where an edge $V_i\rightarrow V_j$ denotes that $V_i$ is a parent (direct cause) of $V_j$ while $V_j$ is a child (direct effect) of $V_i$.
The triplet $\langle V, G, P(V)\rangle$ is called a BN if and only if $\langle V, G, P(V)\rangle$ satisfies the Markov condition: every node of $G$ is independent of any subset of its non-descendants conditioning on the parents of the node~\citep{pearl2014probabilistic}.
A BN encodes the joint probability $P(V)$ over $V$ and decomposes $P(V)$ into a product of the conditional probability distributions of the variables given their parents in $G$. Let $Pa(V_i)$ be the set of parents of $V_i$ in $G$.  Then,  we have
\begin{equation}\label{eq1}
P(V_1, V_2,\cdots,V_{M+1}) = \prod^{M+1}_{i=1}{P(V_i|Pa(V_i))}
\end{equation}


In the following, we introduce the key concepts and assumptions related to BN, Markov boundary, and causality-based feature selection.

\begin{definition}
[Faithfulness]~\citep{pearl2014probabilistic} Given a BN $<V, G, P(V)>$,  $G$ is faithful to $P(V)$ if and only if every conditional independence present in $P$ is entailed by $G$ and the Markov condition. $P(V)$ is faithful to $G$ if and only if there exists a DAG $G$ such that $G$ is faithful to $P(V)$.
\label{def2-4}
\end{definition}

\begin{definition}
[Causal sufficiency]~\citep{pearl2014probabilistic,spirtes2000causation} Causal sufficiency assumes that any common cause of two or more variables in $V$ is also in $V$.
\label{def2-yk1}
\end{definition}




Before introducing Markov boundary, we first present the concept of Markov blanket. 

\begin{definition}[Markov blanket, Mb]~\citep{pearl2014probabilistic} 
A Markov blanket of the class variable $C$ ($Mb(C)$) in $V$ is a set of variables conditioned on which all other variables are independent of $C$, that is, for every $V_i\in V\setminus (Mb(C)\cup C)$, $C\indep V_i|Mb(C)$.
\label{def2-61}
\end{definition}

Clearly, a variable may have multiple Markov blankets. For example, the set of all variables $V$ excluding $C$ is also a Markov blanket of $C$. In practice, we are often interested in minimal Markov blankets.

\begin{definition}[Markov boundary, MB]~\citep{pearl2014probabilistic} 
If no proper subset of $Mb(C)$ satisfies the definition of Markov blanket of $C$, then $Mb(C)$ is called the Markov boundary of $C$, denoted as $MB(C)$.
\label{def2-62}
\end{definition}

Theorem~\ref{the2-1} states the uniqueness of MBs and what the MB of a node is in a BN.

\begin{theorem}~\citep{pearl2014probabilistic}
Under the faithfulness assumption, the MB of a node in a BN is unique and  it consists of the node's parents (direct causes), children (direct effects), and spouses (other parents of the node's children).
\label{the2-1}
\end{theorem}

In a BN, the MB of a node renders the node statistically independent of all the remaining nodes conditioning on the MB~\citep{pearl2014probabilistic}, as shown in Proposition~\ref{pro2-0} below.
\begin{proposition}~\citep{pearl2014probabilistic}
In a BN, let $MB(X)$ be the MB of node $X$,  $\forall Y\in V\setminus (MB(X)\cup X)$, $X\indep Y|MB(X)$ holds.
\label{pro2-0}
\end{proposition}
For example, given the MB of \emph{Lung cancer} in Figure~\ref{fig2-1}, i.e., \emph{Smoking}, \emph{Gentics}, \emph{Coughing}, \emph{Fatigue}, and \emph{Allergy}, \emph{Lung cancer} is independent of the remaining nodes.
Proposition~\ref{pro2-0} illustrates that learning the MB of the class variable is actually a procedure of feature selection~\citep{tsamardinos2003towards,aliferis2010local1}.
Koller and Sahami~\citep{koller1996toward} were the first to introduce the concept of MBs to feature selection. The work in~\citep{tsamardinos2003towards,yu2018unified} stated that under the faithfulness assumption, (1) the strongly relevant features belong to the MB of  the class variable, and (2) the MB is the minimal feature subset with maximum predictivity for classification. Existing  causality-based feature selection algorithms aim to learn the MB of the class variable or a subset of the MB~\citep{guyon2007causal,aliferis2010local1}.


\begin{algorithm}[t]\label{alg2-1}
\caption{Standard forward-backward selection (SFBS)}
\begin{algorithmic}[1]

\STATE \textbf{Input}: Feature Set $F$ and the class variable $C$\\
 \textbf{Output}: The set of selected $S$

\STATE $S=\emptyset$;
\STATE //Forward phase: Adding features to  $S$
\REPEAT
\STATE Identify the most informative feature $X\in F$ by selection criterion $\Phi$;
\IF {$\Phi (S\cup X)>\Phi (S)$}
       \STATE $S=S\cup X$ and $F=F\setminus X$;
\ENDIF
\UNTIL{no features in $F$ are added to $S$};

\STATE //Backward phase: Removing features from $S$
\REPEAT
\STATE Find the least informative feature $Y\in S$ by selection criterion $\Phi$;
\IF {$\Phi (S\setminus Y)\geq\Phi (S)$}
      \STATE $S=S\setminus Y$;
\ENDIF
\UNTIL{no features in $S$ are removed}
\STATE Output $S$
\end{algorithmic}
\end{algorithm}

\begin{algorithm}[t]\label{alg2-2}
\caption{Interleaving forward-backward selection (IFBS)}
\begin{algorithmic}[1]

\STATE \textbf{Input}: Feature Set $F$ and the class variable $C$\\
 \textbf{Output}: The set of selected $S$

\STATE $S=\emptyset$;
\REPEAT
\STATE //Forward phase: Adding features to  $S$
\STATE Identify the most informative feature $X\in F$ by selection criterion $\Phi$;
\IF {$\Phi (S\cup X)>\Phi (S)$}
       \STATE $S=S\cup X$ and $F=F\setminus X$;
       \STATE //Backward phase: Removing features from $S$
        \REPEAT
        \STATE Find the least informative feature $Y\in S$ by selection criterion $\Phi$;
        \IF {$\Phi (S\setminus Y)\geq\Phi (S)$}
      \STATE $S=S\setminus Y$;
\ENDIF
        \UNTIL{no features in $S$ are not removed}
\ENDIF
\UNTIL{no features in $F$ are added to $S$};
\STATE Output $S$
\end{algorithmic}
\end{algorithm}

\subsection{The general strategy of causality-based feature selection}

Forward-backward feature selection is one of the most basic and commonly-used feature selection frameworks.
The forward phase of forward-backward selection starts with a (usually empty) set of features and adds features to it, until a given stopping criterion is met, a the backward phase of forward-backward selection starts with a set of features (usually obtained from the forward phase) and then removes features from that set until a stopping criterion is met. Under the forward-backward framework, there are two general strategies for feature selection. The standard forward-backward feature selection (SFBS) strategy (Algorithm 1) starts with a forward phase for selecting a subset of candidate features $S$ and then a backward phase  for removing false positives from $S$. The interleaving forward-backward feature selection (IFBS) strategy (Algorithm 2) performs the froward phase and backward phase alternatively. Specifically, if there are new features added to $S$ at the forward phase, IFBS immediately triggers the backward phase and implements both phases alternatively. Existing causality-based feature selection algorithms adopt either the SFBS or the IFBS strategy, and they employ a selection criterion $\Phi$, such as information gain, independence tests, and score criteria, to add/remove features to/from $S$.


\begin{table}[t]
\centering
\caption{Representative constraint-based algorithms using conditional independence tests}
\begin{tabular}{|l|l|}
\hline
Category                                        & Representative algorithm                             \\ \hline
\multirow{9}{*}{\begin{tabular}[c]{@{}l@{}}Simultaneous MB learning\\(learning PC and spouses simultaneously\\ and do not distinguish PC from spouses)\end{tabular}}   
                                                & GSMB~\citep{margaritis2000bayesian}    \\ \cline{2-2}
                                                & IAMB~\citep{tsamardinos2003towards}      \\ \cline{2-2}
                                                & IAMBnPC~\citep{tsamardinos2003algorithms}   \\ \cline{2-2}
                                                &IAMB-IP~\citep{pocock2012informative}   \\ \cline{2-2}
                                                & Fast-IAMB~\citep{yaramakala2005speculative}    \\ \cline{2-2}
                                                & Inter-IAMB~\citep{tsamardinos2003algorithms}   \\ \cline{2-2}
                                                & Inter-IAMBnPC~\citep{tsamardinos2003algorithms}     \\ \cline{2-2}
                                                &FBED$^K$~\citep{borboudakis2019forward}  \\ \cline{2-2}
                                                &PFBP~\citep{tsamardinos2019greedy}\\ \hline
\multirow{8}{*}{\begin{tabular}[c]{@{}l@{}}Dvide-and-conquer MB learning\\(learning PC and spouses separately)\end{tabular}} 
                                                & MMMB~\citep{tsamardinos2003time}         \\ \cline{2-2}
                                                & HITON-MB~\citep{aliferis2003hiton}           \\ \cline{2-2}
                                                & Semi-HITON-MB~\citep{aliferis2010local1}         \\ \cline{2-2}
                                                & PCMB~\citep{pena2007towards}             \\ \cline{2-2}
                                                & IPCMB~\citep{fu2008fast}                 \\ \cline{2-2}
                                                & MBOR~\citep{de2008novel}                  \\ \cline{2-2}
                                                & STMB~\citep{gao2017efficient}          \\ \cline{2-2}
                                                & CCMB~\citep{wu2019ccmb}\\ \hline
\multirow{2}{*}{\begin{tabular}[c]{@{}l@{}}MB learning with interleaving PC and \\spouse learning\end{tabular}} 
                                                & BAMB~\citep{ling2019BAMB}      \\ \cline{2-2}
                                                & EEMB~\citep{ling2020EEMB}   \\ \hline
\multirow{6}{*}{\begin{tabular}[c]{@{}l@{}}MB learning with relaxed assumptions\\ (e.g. the faithfulness assumption or \\causal sufficiency assumption)\end{tabular}}   
                                                 & KIAMB~\citep{pena2007towards}            \\ \cline{2-2}
                                                & TIE*~\citep{statnikov2013algorithms}     \\ \cline{2-2}
                                                & SGAI~\citep{yu2017markov}      \\ \cline{2-2}
                                                & LCMB~\citep{liu2016swamping}          \\ \cline{2-2}
                                                & WLCMB~\citep{liu2016swamping}            \\ \cline{2-2}
                                                &M3B~\citep{yu2018mining}               \\ \hline
\multirow{5}{*}{\begin{tabular}[c]{@{}l@{}}MB learning with special purpose\\(e.g. multiple data sets, distribution\\ shift, and weak supervision) \end{tabular}}
                                               &MIMB~\citep{yu2018discovering}     \\ \cline{2-2}
                                               &MCFS~\citep{yu2019MCFS}    \\ \cline{2-2}
                                               & MIAMB and MKIAMB~\citep{liu2018markov}    \\ \cline{2-2}
                                               &BASSUM~\citep{cai2011bassum}   \\ \cline{2-2}
                                               &Semi-IAMB~\citep{sechidis2018simple}   \\ \hline
\end{tabular}
\label{tb3-1}
\end{table}

\section{Constraint-based methods}\label{sec3}

In this section, we will discuss the constraint-based MB learning methods to learn the MB or PC of the class variable, i.e., the methods using conditional independence tests.  Constraint-based methods can be categorized into five types: simultaneous MB learning, divide-and-conquer MB learning,  MB learning with interleaving PC and spouse learning, MB learning  with relaxed assumptions, and MB learning with special purpose. A summary of the representative algorithms of the five types is given in Table~\ref{tb3-1}.

In the following, Section~\ref{sec32} presents the basis of constraint-based methods. Section~\ref{sec31} gives the brief discussions of the five types of constraint-based methods.  Section~\ref{sec33} extensively reviews existing constraint-based methods of each type.

\subsection{Basis of the constraint-based methods}\label{sec32}

The constraint-based methods are mainly based on Propositions~\ref{pro2-1} and~\ref{pro2-2} below.
Proposition~\ref{pro2-1} illustrates the dependent relations between a node and its parents (or children). It states that if $V_i$ is a parent or a child of  $V_j$ in a BN, $V_i$ and $V_j$ are not independent conditioning on any subsets of $V\setminus\{V_i,V_j\}$.  For example, in Figure~\ref{fig2-1}, there is an edge between \emph{Smoking}  and \emph{Lung cancer}, and thus \emph{Smoking} and \emph{Lung cancer} are not independent given any subsets of  the remaining features. 


\begin{proposition}\citep{spirtes2000causation}\label{pro2-1}
In a BN, if node $V_i$ is a parent (or a child) of $V_j$,  then $\forall S\subseteq V\setminus \{V_i, V_j\}$, $V_i\nindep V_j|S$.
\label{pro2-1}
\end{proposition}

Proposition~\ref{pro2-2} presents the relation between a node and its spouses in a BN. It indicates that if $V_i$ is a spouse of $V_k$ and $V_j$ is their common child, there exists a subset $S\subseteq V\setminus\{V_i,V_j,V_k\}$ such that $V_i$ and $V_k$ are independent given $S$ but they are dependent given $S\cup V_j$. For instance, \emph{Allergy} is the spouse of \emph{Lung cancer}  in  Figure~\ref{fig2-1}. \emph{Allergy} and \emph{Lung cancer}  are independent ($S$ is an empty set), but they are dependent conditioning on their common child \emph{Coughing}. Proposition~\ref{pro2-2} shows that  \emph{Allergy} (spouse) and  \emph{Coughing} together carry more predictive information about  \emph{Lung cancer} than \emph{Coughing} only.  Proposition~\ref{pro2-2} also states that spouses of $V_k$ consist of all parents of the children of $V_k$ (excluding $V_k$).

\begin{proposition}\citep{spirtes2000causation}\label{pro2-2}
In a BN, assuming that $V_i$ is adjacent to $V_j$, $V_j$ is adjacent to $V_k$, and $V_i$ is not adjacent to $V_k$ (e.g., $V_i\rightarrow V_j\leftarrow V_k$), if
$\exists S\subseteq V\setminus\{V_i, V_j, V_k\}$ such that $V_i\indep V_k|S$ and $V_i\nindep V_k|\{S, V_j\}$ hold, $V_i$ is a spouse of $V_k$.
\label{pro2-2}
\end{proposition}

Existing constraint-based methods adopt either the SFBS strategy or the IFBS strategy, and they employ the statistical independence tests as the selection criteria, denoted as $\Phi$, to add/remove features to/from $S$ as follows.

Given the class variable $C$, in SFBS (or IFBS), at each iteration,  let $S$ be a set of features currently selected, if $X\in V\setminus\{C\cup S\}$ and $C$ are conditionally independent conditioning on $S$ (or a subset $S'\subset S$), that is, $X\indep C|S$ (or $X\indep C|S'$), $X$  does not provide any predictive information to $C$ conditioning on $S$ (i.e., $\Phi (S\cup X)\leq\Phi (S)$ in SFBS or IFBS). In this case, $X$ is not to be added to $S$ at the forward phase or removed from $S$ at the backward phase.

Five types of conditional independence tests are used by current constraint-based methods, $\lambda^2$ test, $G^2$ test, mutual information for discrete features~\citep{mcdonald2009handbook}, Fisher’s Z tests for continuous features with linear relations with additive Gaussian errors~\citep{pena2008learning}, and kernel-based tests for continuous features with nonlinearity and non-Gaussian noise~\citep{zhang2012kernel}.

\subsection{Overview of constraint-based methods}\label{sec31}
 In the section, we will give a brief overview of the five types of constraint-based methods. The detailed review of the representative algorithms of each type will be presented in Section~\ref{sec33}.

\begin{figure}[t]
\centering
\includegraphics[height=0.8in,width=4in]{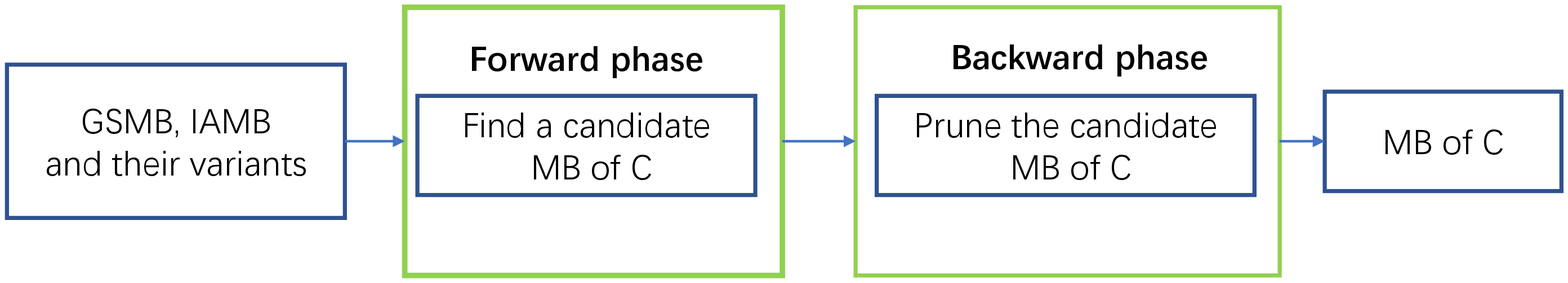}
\caption{Simultaneous MB learning}
\label{fig311}
\end{figure}

\textbf{1. Simultaneous MB learning.}
Given the class variable $C$, a simultaneous MB learning algorithm aims to find parents, children, and spouses of $C$ simultaneously, and do not distinguish PC (parents and children) of $C$ from its spouses during the MB learning.
As shown in Figure~\ref{fig311}, the simultaneous MB learning approach adopts a forward-backward strategy to greedily learn a MB of $C$ by conditioning on the entire candidate MB of $C$ ($CMB(C)$) currently selected at each iteration.
The representative simultaneous MB learning algorithms include GSMB~\citep{margaritis2000bayesian}, IAMB~ \citep{tsamardinos2003towards}, IAMBnPC~\citep{tsamardinos2003algorithms}, Fast-IAMB~\citep{yaramakala2005speculative}, Inter-IAMB~\citep{tsamardinos2003algorithms}, Inter-IAMBnPC~\citep{tsamardinos2003algorithms}, IAMB-IP~\citep{pocock2012informative}, FBED$^K$~\citep{borboudakis2019forward}, and PFBP~\citep{tsamardinos2019greedy}.
The GSMB algorithm was the first algorithm for learning a MB of the class variable without learning an entire Bayesian network. IAMB and its variants are all the improved versions of GSMB.  Inter-IAMB interleaves the forward phase and the backward phase of IAMB. Both FBED$^K$ and PFBP are the state-of-the-art variants of IAMB.

Due to  the use of the entire $CMB(C)$ currently selected for conditional independence tests at each computation, existing simultaneous MB learning algorithms reduce the number of independence tests, but require more data samples for each test since the number of data samples required is exponential to the size of the conditioning set. Thus the simultaneous MB learning algorithms are time efficient but not data efficient. When the sample size of a data set is not big enough, these algorithms cannot find the MB accurately. The quality of learnt MBs of these algorithms degrades greatly in practical settings due to the limited number of samples. They are expected to perform the best in problems where $MB(C)$ is small.

\textbf{2. Divide-and-conquer MB learning.}
The divide-and-conquer MB learning approach aims to reduce the data requirements of the simultaneous MB learning approach. This approach  breaks the problem of learning $MB(C)$ into two subproblems: first, learning parents and children of $C$ (i.e., $PC(C)$), and second, learning the spouses of $C$ (i.e., $SP(C)$).
As for learning $PC(C)$, the divide-and-conquer approach does not use the entire $PC(C)$ as the conditioning set for conditional independence tests when determining whether feature $X$ is a candidate member of $PC(C)$.  
Instead, it makes use of the subsets of $PC(C)$  which is much smaller than $CMB(C)$ used by the simultaneous MB learning approach when making decisions. Therefore, the divide-and-conquer MB learning approach needs significantly smaller number of samples than the simultaneous MB learning approach.
For instance, to determine whether feature $X$ is a candidate member of $PC(C)$, the divide-and-conquer approach explores possible subsets of $PC(C)$. If there exists a subset $S\subseteq PC(C)$ such that $X$ and $C$ are conditional independent given $S$, this subset exploring process will terminate and $X$ will be discarded and never considered again. However, for the simultaneous MB learning approach,  such as GSMB, IAMB, and their variants, the discarded features will be reconsidered many times (for identifying spouses).

\begin{figure}[t]
\centering
\includegraphics[height=2.7in,width=4.5in]{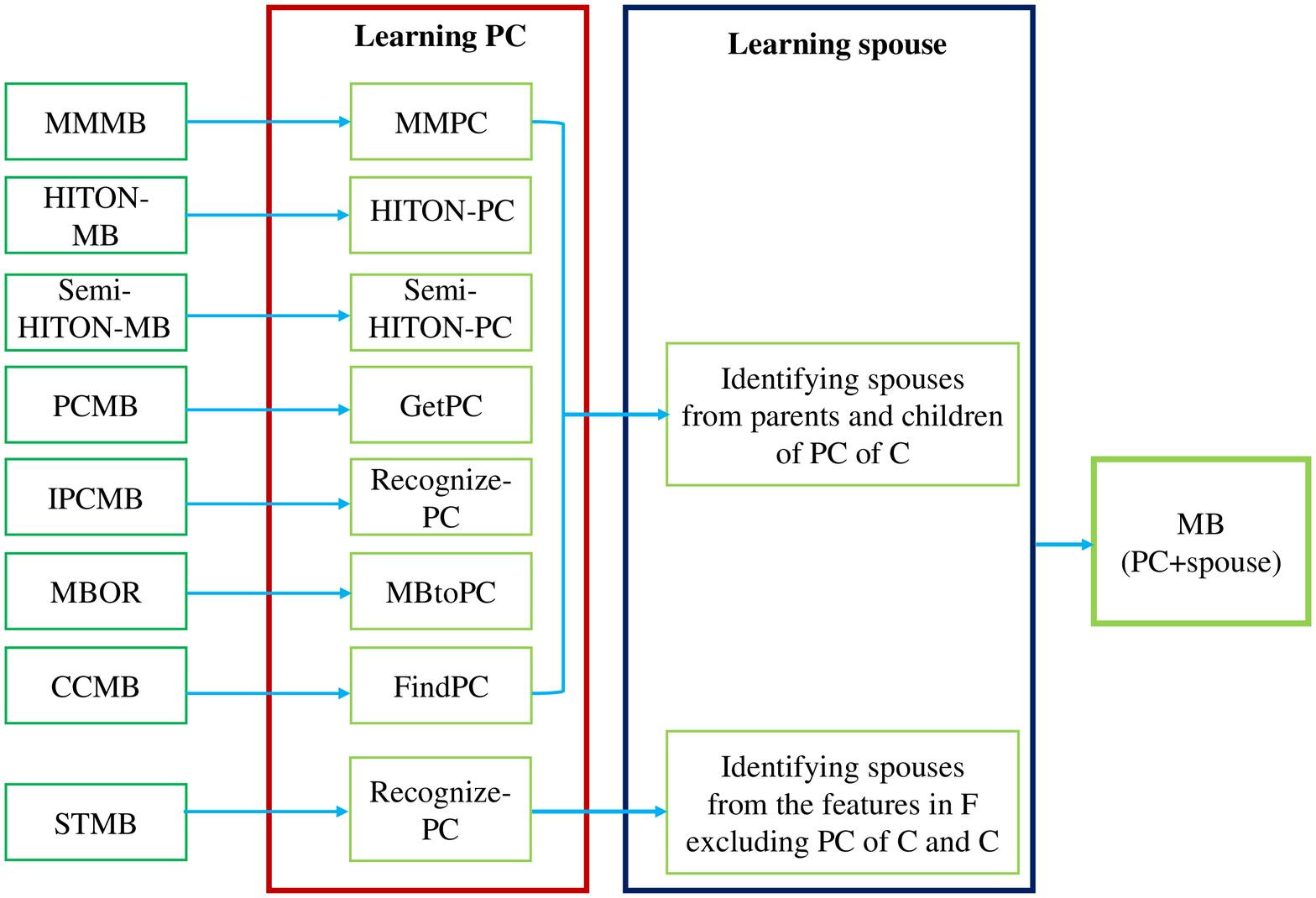}
\caption{Divide-and-conquer MB learning}
\label{fig312}
\end{figure}

The representative divide-and-conquer algorithms include MMMB~\citep{tsamardinos2003time}, HITON-MB~\citep{aliferis2003hiton}, semi-HITON-MB~\citep{aliferis2003hiton}, PCMB~\citep{pena2005scalable}, IPCMB~\citep{fu2008fast}, MBOR~\citep{de2008novel}, STMB~\citep{gao2017efficient}, and CCMB~\citep{wu2019ccmb}.
The main differences between those algorithms lie in the strategies of identifying parents and children of $C$ and the strategies of finding spouses of $C$, as shown in Figure~\ref{fig312}. This figure also presents the general steps of existing divide-and-conquer MB learning algorithms (i.e., learning PC and identifying spouses separately), and the PC learning used by the eight representative MB algorithms respectively.

The divide-and-conquer MB learning methods are data efficient but not time efficient.  Although they mitigate the problem of the large sample requirement, existing divide-and-conquer MB learning algorithms will be computationally expensive when the size of currently selected features becomes large.

\begin{figure}[t]
\centering
\includegraphics[height=1.8in,width=4.5in]{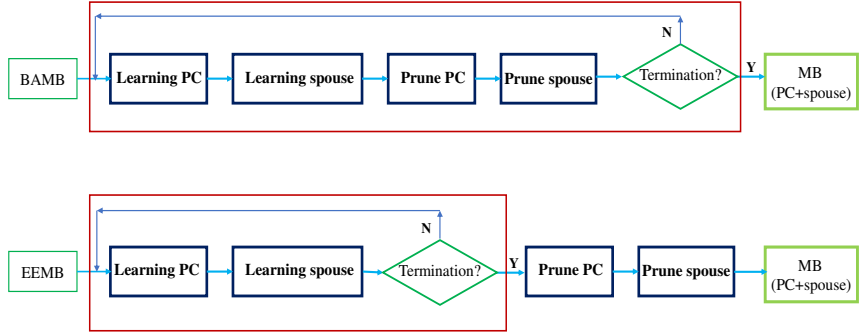}
\caption{MB learning with interleaving PC and spouse learning}
\label{fig313}
\end{figure}

\textbf{3. MB learning with interleaving PC and spouse learning.}
This approach  is an extension of the divide-and-conquer approach.  Instead of learning PC and identifying spouses separately, this approach implements the  PC leaning phase and the spouse identifying phase alternatively. Specifically, once a candidate member of PC of $C$ is added to the candidate $PC(C)$ at the PC learning phase, this approach triggers the spouse learning phase immediately.  The representative algorithms include BAMB~\citep{ling2019BAMB} and EEMB~\citep{ling2020EEMB}. The difference between BAMB and EEMB is shown in Figure~\ref{fig313}, where we can see that BAMB learns the candidate PC and spouse sets of $C$ and removes false positives from the two candidate sets in one go, while EEMB breaks BAMB into two independent subroutines: learning  and pruning. 

By interleaving PC and spouse learning, BAMB and EEMB attempt to keep both candidate PC and spouse sets as small as possible for achieving the trade-off between data efficiency and time efficiency. However, due to false PC inclusions, many false spouses may enter the candidate spouse set, leading to a large size of the candidate spouse set, which will degrade the performance of BAMB and EEMB.

\textbf{4. MB learning  with relaxed assumptions.}
The above algorithms are designed  to learn the MB of the class variable under the faithfulness and causal sufficiency assumptions. In fact, both assumptions are often violated in practice. 

The theoretical result has stated that if a data distribution satisfies the faithfulness assumption, the MB of the class variable is unique~\citep{pearl2014probabilistic}. When the faithfulness assumption is violated,  the MB of learnt from the data may not be unique~\citep{pena2007towards,statnikov2013algorithms}.
To deal with the violation of the faithfulness assumption, some research work has been done for identifying multiple MBs without the assumption, such as KIAMB~\citep{pena2007towards}, TIE*~\citep{statnikov2013algorithms}, SGAI~\citep{yu2017markov},  LCMB~\citep{liu2016swamping}, and WLCMB~\citep{liu2016swamping}.
KIAMB was the first attempt to learn multiple MBs, but it needs to run multiple times and cannot guarantee finding all possible MBs of the class variable. TIE* can find all MBs of the class variable in a data set, but it is computationally expensive, especially when the number of MBs is large. SGAI may be more efficient than TIE* but it is not guaranteed to find all possible MBs of the class variable. WLCMB is motivated by KIAMB and thus it still suffers from the drawbacks of KIAMB. Thus it is still a challenging problem to tackle multiple MB learning when the faithfulness assumption is violated.

When the causal sufficiency assumption does not hold in a data set,  if we still use a MB learning algorithm that assumes causal sufficiency, the MB learnt from the data set may not properly indicate the true causal relations.
Yu et al.~\citep{yu2018mining}  proposed the M3B algorithm to deal with the situation when the assumption of causal sufficiency is violated.
However, M3B uses a backward strategy to learn PC of the class variable and needs to perform an exhaustive search over the currently selected PC set. Thus it also suffers from time efficiency and incorrect test problems of the constraint-based MB learning in general.

\textbf{5. MB learning with special purpose.}
Beyond the algorithms discussed above, several MB learning algorithms have been proposed for special purposes, including the MIMB algorithm for identifying a MB of a class variable from multiple data sets~\citep{yu2018discovering}, the MCFS algorithm for stable prediction with distribution shift~\citep{yu2019MCFS}, the MIAMB and MKIAMB algorithms for learning a MB of multiple class variables~\citep{liu2018markov}, and the BASSUM and Semi-IAMB algorithms for MB learning with weak supervision~\citep{cai2011bassum,sechidis2018simple}. 
These studies have shown that causal properties of features can facilitate semi-supervised learning and feature selection with distribution shifts. Moreover the intersection of machine learning and causal discovery has attracted increasing attention in areas beyond feature selection. For example, causal knowledge has inspired efficient transfer learning and domain adaptation methods for accurate prediction across different domains~\citep{RojSchTurPet18,magliacane2018domain}. It is a promising research area to link machine learning research with causality to develop explainable and robust machine learning methods and solutions to causal discovery for data analytics.

\subsection{Detailed review of constraint-based methods}\label{sec33}

\subsubsection{Methods of simultaneous MB learning}\label{sec331}
In this subsection, we firstly introduce the methods using SFBS,  including GSMB, IAMB and the two extensions of IAMB, which are IAMBnPC and IAMP-IP. Then we introduce the methods employing IFBS, which are Fast-IAMB, Inter-IAMB and Inter-IAMBnPC. Since FBEDk and PFBP are the state-of-the-art simultaneous MB learning algorithms and the extensions of IAMB, they will be introduced at the end.

\begin{algorithm}[t]\label{alg4-1}
\centering
\caption{The instantiation of SFBS for simultaneous MB learning}
\begin{algorithmic}[1]

\STATE \textbf{Input}: Feature Set $F$ and the class variable $C$\\
 \textbf{Output}: $CMB(C)$

\STATE $CMB(C)=\emptyset$;
\STATE //Forward phase: Adding candidate MB (relevant) features to $CMB(C)$
\REPEAT
\STATE Select a feature $X\in F$;
\IF {$X\nindep C|CMB(C)$}
       \STATE $CMB(C)=CMB(C)\cup X$ and $F=F\setminus X$;
\ENDIF
\UNTIL{no features in $F$ are added to $CMB(C)$};

\STATE //Backward phase: Removing false positives from $CMB(C)$
\REPEAT
\STATE Select a feature $Y\in CMB(C)$;
\IF {$Y\indep C|CMB(C)\setminus Y$}
      \STATE $CMB(C)=CMB(C)\setminus Y$;
\ENDIF
\UNTIL{no features in $CMB(C)$ are removed}
\STATE Output $CMB(C)$
\end{algorithmic}
\end{algorithm}

\textbf{GSMB.} The Growing-Shrinking MB (GSMB) learning algorithm~\citep{margaritis2000bayesian,margaritis2009toward} instantiates the SFBS framework for simultaneous MB learning, as shown in Algorithm 3.  Let $CMB(C)$ be the candidate MB of $C$ currently selected, in the forward (growing) phase (Steps 5 to 10 in Algorithm 3), at each iteration,  if $\exists X\in F\setminus CMB(C)$ such that $X\nindep C|CMB(C)$ holds, GSMB adds $X$ to $CMB(C)$, until no features within $F\setminus CMB(C)$ are added to $CMB(C)$. In the backward (shrinking) phase (Steps 12 to 17 in Algorithm 3), GSMB sequentially removes from $CMB(C)$ the false positive $Y\in CMB(C)$ satisfying $Y\indep C|CMB(C)\setminus Y$.
However, at Step 5 of the forward phase, GSMB uses a static heuristic that at each time GSMB randomly selects a feature $X\in F$ satisfying $X\nindep C|CMB(C)$ and adds it to $CMB(C)$. The static heuristic may make many false positives enter $CMB(C)$ in the forward phase, leading to the growing of the size of $CMB(C)$. Given a fixed size of data samples, the larger size of  $CMB(C)$, the more unreliable the independence tests. Thus the heuristic makes GSMB ineffective in coping with a data set of small sample size but high dimensionality.


\textbf{IAMB, IAMBnPC, and IAMB-IP.} To tackle the problem with GSMB, the incremental association Markov boundary (IAMB) algorithm~\citep{tsamardinos2003towards} uses a dynamic heuristic at Step 5 in Algorithm 3 of the forward phase. At each iteration, IAMB adds to $CMB(C)$ the feature $X\in  F\setminus CMB(C)$ with the highest association with $C$ conditioning on the current $CMB(C)$  if $X\nindep C|CMB(C)$ holds. This dynamic heuristic makes the features that belong to $MB(C)$ enter $CMB(C)$ as early as possible and reduces as much as possible the chance of false positives to enter $CMB(C)$ during the forward phase.
Accordingly, IAMB performs better (with lower time complexity and lower data sample requirement) than GSMB since fewer false positives will be added to $CMB(C)$ in the forward phase.
However, the number of required data samples of IAMB is still exponential with the size of $CMB(C)$ since the size of $CMB(C)$ may become large in the forward phase.
To mitigate this problem, several variants of IAMB were proposed, such as IAMBnPC~\citep{tsamardinos2003algorithms},  Inter-IAMB~\citep{tsamardinos2003algorithms}, inter-IAMBnPC~\citep{tsamardinos2003algorithms}, and Fast-IAMB~\citep{yaramakala2005speculative}. Compared to IAMB, IAMBnPC only substitutes the backward phase (Steps 11 to 16 in Algorithm 3)  as implemented in IAMB with the PC algorithm~\citep{spirtes2000causation}. IAMBnPC is more data efficient (with lower data sample requirement) than IAMB since the PC algorithm runs only on the subsets of the current $CMB(C)$ instead of conditioning on the entire $CMB(C)$.
To leverage prior knowledge, the IAMB-IP (IAMB-Informative Prior) algorithm was proposed in~\citep{pocock2012informative}. It can  incorporate domain knowledge priors and structure sparsity priors to improve the performance of MB learning when the data set is of small sample size but  high dimensionality.

\begin{algorithm}[t]\label{alg4-2}
\centering
\caption{The instantiation of IFBS for simultaneous MB Learning}
\begin{algorithmic}[1]

\STATE \textbf{Input}: Feature Set $F$ and the class variable $C$\\
 \textbf{Output}: $CMB(C)$

\STATE $CMB(C)=\emptyset$;
\REPEAT
\STATE //Forward phase: Adding candidate MB (relevant) features to $CMB(C)$
\STATE Select a feature $X\in F$ with the highest association with $C$;
\IF {$X\nindep C|CMB(C)$}
       \STATE $CMB(C)=CMB(C)\cup X$ and $F=F\setminus X$;
       \STATE //Backward phase: Removing false positives from $CMB(C)$
       \REPEAT
          \STATE Select a feature $Y\in CMB(C)$;
          \IF {$Y\indep C|CMB(C)\setminus Y$}
             \STATE $CMB(C)=CMB(C)\setminus Y$;
          \ENDIF
\UNTIL{no features in $CMB(C)$ are removed}
\ENDIF
\UNTIL{no features in $F$ are added to $CMB(C)$};
\STATE Output $CMB(C)$
\end{algorithmic}
\end{algorithm}

\textbf{Inter-IAMB and Inter-IAMBnPC.} 
These two algorithms adopt the IFBS framework, which is the key difference between them and IAMB. Algorithm 4 shows how they instantiate IFBS for simultaneous MB learning. The goal of the interleaving is to keep the size of $CMB(C)$ as small as possible during all steps of the algorithms' execution. Comparing to Inter-IAMB,  the Inter-IAMBnPC algorithm substitutes the backward phase as implemented in inter-IAMB with the PC algorithm (Steps 9 to 14 in Algorithm 4).

\textbf{Fast-IAMB.}  Similar to Inter-IAMB and Inter-IAMBnPC, Fast-IAMB instantiates IFBS as shown in Algorithm 4.
However, different from IAMB and its other variants discussed above, Fast-IAMB adopts an aggressively greedy strategy in the forward phase to make it more efficient. Specifically,  at Steps 5 to 6 in Algorithm 4, Fast-IAMB does not add one feature to $CMB(C)$ then immediately triggers the backward phase. Instead, Fast-IAMB greedily adds as many features conditionally dependent on $C$ given the current $CMB(C)$ as possible in the forward phase until a conditional independence test is not reliable (i.e., we do not have enough data for conducting the test).
When a test is not reliable in the forward phase, the backward phase is triggered.
A reliable independence test for $C$ and $X\in F\setminus CMB(C)$ given $CMB(C)$  should satisfy the rule that  the average number of instances per cell of the contingency table of $X\cup C\cup CMB(C)$ must be at least $k$, i.e., $N/\{r_X*r_C*r_{CMB(C)}\}\geq k$ where the  minimum value of $k$  is set to 5 for reliable tests as suggested by Agresti~\citep{agresti2011categorical}, $N$ is the total number of data samples, and $r_C$ denotes the number of discrete values that $C$ takes.
By the rule, at Steps 5 to 6 in Algorithm 4, Fast-IAMB will not perform a test when it is not reliable. This checking not only speeds up Fast-IAMB, but also reduces the risk of unreliable independence tests.

\textbf{FBED$^K$.} Borboudakis and Tsamardinos~\citep{borboudakis2019forward} generalized the IAMB framework for feature selection and proposed the FBED$^K$ (Forward-Backward selection with Early Dropping) algorithm to speed up IAMB. 
For IAMB, in the forward phase,  at each iteration, it should reconsider all remaining features (including all discarded features at each iteration) to find the next best candidate. 
To tackle the issues, $FBED^K$ adopts an early dropping strategy in the forward phase. The main idea  is that at each forward iteration, $FBED^K$ removes the features that are conditionally independent of $C$ given the current $CMB(C)$ from the remaining features in $F$ instead of keeping them in $F$. This leads to quickly reduce the number of candidate features in $F$, while keeping relevant features in it. A run of  the forward phase with the early dropping terminates until $F$ is empty. Then the forward phase is allowed to run up to $K$ additional times to reconsider features dropped previously until no features can be dropped. Finally, the backward phase is applied to $CMB(C)$ obtained at the forward phase, and this is the same as the backward phase of IAMB.
$FBED^K$ significantly improves computational efficiency, while retaining competitive accuracy.

\textbf{PFBP.}  Motivated by FBED$^K$,  the Parallel Forward-Backward with Pruning (PFBP) algorithm was proposed for improving IAMB to tackle big data with high dimensionality~\citep{tsamardinos2019greedy}.
PFBP enables computations to be performed in a parallel way by partitioning data both in terms of rows (samples) as well as columns (features) and using meta-analysis techniques to combine results of local computations.

In addition to the early dropping strategy proposed in~\citep{borboudakis2019forward}, PFBP also proposed two new heuristics of early stopping with the consideration of features within the same iteration and early returning the current best feature for addition or removal.
It has been shown that PFBP can scale to millions of features and millions of training samples, and achieves a super-linear speedup with increasing sample size and linear scalability with respect to the number of features and processing cores. 

\subsubsection{Methods of divide-and-conquer MB learning}\label{sec332}
In this subsection, we will  discuss eight representative  divide-and-conquer algorithms, i.e., MMMB~\citep{tsamardinos2003time}, HITON-MB~\citep{aliferis2003hiton}, semi-HITON-MB~\citep{aliferis2003hiton}, PCMB~\citep{pena2005scalable}, IPCMB~\citep{fu2008fast}, MBOR~\citep{de2008novel}, STMB~\citep{gao2017efficient}, and CCMB~\citep{wu2019ccmb}.
As illustrated in Figure~\ref{fig312}, given the class variable $C$, how to learn its parents and children and identify its spouses is the main difference between those algorithms.
Generally speaking, there are three strategies for learning parents and children of $C$: SFBS, IFBS and the backward framework. SFBS and IFBS for PC learning are very similar to those for MB learning.  The instantiations of SFBS and IFBS for PC learning are present in Algorithms 5 and 6 respectively, while the backward  framework for PC learning is shown in Algorithm 7.

\begin{algorithm}[t]\label{alg4-3}
\centering
\caption{The instantiation of SFBS for PC Learning}
\begin{algorithmic}[1]

\STATE \textbf{Input}: Feature set $F$ and the class variable $C$\\
 \textbf{Output}: $CPC(C)$
\STATE $CPC(C)=\emptyset$;
\STATE // Filtering out irrelevant features by Proposition~\ref{pro2-1} (if $X\indep C$  holds, $X\notin PC(C)$)
\STATE $R=F\setminus S'$  ($\forall X\in S',\ X\indep C|\emptyset$);
\STATE //Forward phase: Adding candidate PC (or relevant) features  to $CPC(C)$
\REPEAT
\STATE Select the best feature $X\in R$ with a greedy strategy;
\STATE $CPC(C)=CPC(C)\cup \{X\}$; $R=R\setminus X$;
\UNTIL{no features in $R$ are added to $CPC(C)$};

\STATE //Backward phase: Removing false positives from $CPC(C)$;
\REPEAT
\STATE Consider each feature $Y\in CPC(C)$;
\IF {$\exists S\subseteq CPC(C)\setminus Y$ s.t. $Y\indep C|S$}
       \STATE $CPC(C)=CPC(C)\setminus Y$;
\ENDIF
\UNTIL{no features in $CPC(C)$ are removed};
\STATE Output $CPC(C)$
\end{algorithmic}
\end{algorithm}

\begin{algorithm}[t]\label{alg4-4}
\centering
\caption{The instantiation of IFBS for PC Leaning}
\begin{algorithmic}[1]

\STATE \textbf{Input}: Feature set $F$ and the class variable $C$\\
 \textbf{Output}: $CPC(C)$

\STATE $CPC(C)=\emptyset$;
\STATE $R=F\setminus S'$  ($\forall X\in S',\ X\indep C|\emptyset$);
\REPEAT
\STATE //Forward phase: Adding candidate PC (relevant) features to $CPC(C)$
\STATE Select the best feature $X\in R$ with a greedy strategy;
\STATE $CPC(C)=CPC(C)\cup \{X\}$; $R=R\setminus X$;
\STATE //Backward phase: Removing false positives from $CPC(C)$
\REPEAT
\STATE Consider each feature $Y\in CPC(C)$
\IF {$\exists S\subseteq CPC(C)\setminus Y$ s.t. $Y\indep C|S$}
       \STATE $CPC(C)=CPC(C)\setminus Y$;
\ENDIF
\UNTIL{no features in $CPC(C)$ are removed};
\UNTIL{no features in $R$ are added to $CPC(C)$};
\STATE Output $CPC(C)$
\end{algorithmic}
\end{algorithm}

\begin{algorithm}[t]\label{alg4-5}
\centering
\caption{The Backward Framework for PC Learning}
\begin{algorithmic}[1]

\STATE \textbf{Input}: Feature Set $F$ and the class variable $C$\\
 \textbf{Output}: $CPC(C)$

\STATE $CPC(C)=\{F\}$;
\STATE i=0;
\REPEAT
\FOR {each feature $X$ in $CPC(C)$}
         \IF {$\exists S\subseteq CPC(C)\setminus X$ and $|S|=i$ such that $X\indep C|S$}
                      \STATE $CPC(C)=CPC(C)\setminus X$;
 \ENDIF
\ENDFOR
\STATE i=i+1;
\UNTIL{$i>|CPC(C)|$};
\STATE Output $CPC(C)$
\end{algorithmic}
\end{algorithm}


\textbf{MMMB.}
The MMMB (Max-Min MB) algorithm~\citep{tsamardinos2003time} first employs the MMPC (Max-Min Parents and Children) algorithm to find a candidate set of parents and children of $C$. MMPC~\citep{tsamardinos2003time} utilizes the SFBS framework to search for candidate parents and children of $C$ first, called $CPC(C)$, then prunes $CPC(C)$ at the backward phase, as shown in Algorithm 5.
The novelty of MMPC lies the fact that at Step 7 of the forward phase in Algorithm 5, MMPC proposes a Max-Min Parents and Children (MMPC) greedy search strategy to identify the best feature from $F\setminus CPC(C)$ at each iteration.

Specifically, in the forward phase, at each iteration, given the current $CPC(C)$ (initially $CPC(C)$ is empty), for each feature $X$ in the remaining candidate features (i.e., $X\in F\setminus CPC(C)$),  MMPC first calculates the associations of $X$ and $C$ conditioning on all possible subsets of $CPC(C)$ respectively, and chooses the minimum association as the association of $X$ and $C$.  
Then MMPC chooses the next feature to be included in $CPC(C)$ as the one that exhibits the maximum association among the features in $F\setminus CPC(C)$ and is dependent on $C$, while the features independent of $C$ are discarded and never considered as candidate PC again.  The forward phase terminates until each feature in $F\setminus CPC(C)$  and $C$ are independent given any subsets of $CPC(C)$.
At the backward phase, MMPC examines whether each feature $Y$ in $CPC(C)$ obtained in the forward phase  is independent of $C$ conditioning on all possible subsets of $CPC(C)\setminus Y$. If so, $Y$ is removed from $CPC(C)$; otherwise it is retained.

Now we discuss how to learn spouses of $C$ after $CPC(C)$ is obtained.  The spouses of $C$ are the parents of the children of $C$ excluding $C\cup PC(C)$, i.e., $SP(C)=\bigcup_{Y'\in ch(C)}pa(Y')\setminus (C\cup PC(C))$. However, MMPC cannot distinguish parents from children of $C$ during the procedure of identifying $PC(C)$.  Thus, MMMB considers the union of parents and children of the features in $CPC(C)$ excluding $C\cup CPC(C)$ as the the candidate spouses of $C$, i.e., $SPC(C)=\bigcup_{Y'\in CPC(C)}PC(Y')\setminus(C\cup CPC(C))$.
Then by Proposition~\ref{pro2-2}, for each feature $Y$ in the $SPC(C)$ set and each feature $X$ in $CPC(C)$,  if there exists a subset  $S\subseteq F\setminus\{C,X,Y\}$ ($S$ was identified and stored in the MMPC subroutine) such that both $C\indep Y|S$ and $C\nindep Y|X\cup S$ hold, MMMB considers $Y$ as a spouse of $C$.


\textbf{HITON-MB and  Semi-HITON-MB.}
HITON-MB uses the HITON-PC algorithm to discover $PC(C)$~\citep{aliferis2003hiton}. Different from MMPC, HITON-PC employs the IFBS framework as presented in Algorithm 6.
HITON-PC interleaves the forward phase and the backward phase to make PC learning and false positive removal alternatively. In addition, at Step 6 in Algorithm 6, HITON-PC adopts a simpler search strategy than MMPC for learning candidate parents and children of $C$. Specifically, at Step 6,  at each iteration, HITON-PC removes a feature, called $X$, with the highest association with $C$ conditioning on an empty set from the candidate feature set $R$ and adds it to $CPC(C)$, then triggers the backward phase for removing false positives from the current $CPC(C)$ due to the X's inclusion. 

For identifying the spouses of $C$, in the original version of the HITON-MB algorithm~\citep{aliferis2003hiton}, the idea of HITON-MB is the same as that of MMMB.

However, Pena et al.~\citep{pena2005scalable,pena2007towards} pointed out that MMMB and HITON-MB cannot return the correct MB  even under the faithfulness assumption. They found that  (1) both MMPC and HITON-PC may return a superset of the true PC of $C$, and (2) the spouse discovery procedures of both MMMB and HITON-MB cannot find the correct spouses of $C$.
Tsamardinos et al.~\citep{tsamardinos2006max} also identified the flaw of MMPC in point (1) above independently and proposed a corrected MMPC using the symmetric relation between parents and children in a BN (i.e., symmetric check).  That is,  if $X$ is a parent or a child of $C$, $C$ should be a child or a parent of $X$.  Following this, Aliferis et al.~\citep{aliferis2010local1} proposed a general local learning (GLL) framework and corrected the two flaws  discussed above. In addition, in~\citep{aliferis2010local1}, a new Semi-interleaved HITON-PC (Semi-HITON-PC for brevity) algorithm was proposed to speed up HITON-PC. The difference between Semi-HITON-PC and HITON-PC is that at Step 10 in Algorithm 6, Semi-HITON-PC only considers the elimination of the newly added feature at Step 7 before the candidate feature set $R$ becomes empty and a full feature elimination in $CPC(C)$ will be performed after $R$ is empty.
Employing Semi-HITON-PC,  the semi-HITON-MB algorithm has been proposed accordingly~\citep{aliferis2010local1}.

\textbf{PCMB.}
The parents and children based MB (PCMB) algorithm~\citep{pena2005scalable,pena2007towards} was the first correct divide-and-conquer MB learning algorithm. PCMB uses the two subroutines, called GetPCD and GetPC, to identify $PC(C)$. The GetPCD subroutine is to find $CPC(C)$, and the GetPC subroutine removes false positives in $CPC(C)$ using the symmetric check, i.e., for each feature $X$ in $CPC(C)$, if the set of parents and children of $X$ does not include $C$, $X$ will be removed from $CPC(C)$.

GetPCD adopts the similar idea of MMPC, but the two algorithms have two differences. First, GetPCD adopts the IFBS framework and interleaves the forward and backward phases of MMPC. Second, in the backward phase, for each feature $X$ in the current $CPC(C)$, GetPCD calculates the associations of $X$ and $C$ conditioned on all possible subsets of $CPC(C)$ and chooses the minimum association as the association of $X$ and $C$. If $X$ and $C$ are assessed to be independent given the minimum association,  $X$ will be removed from $CPC(C)$.
Pena et al.~\citep{pena2005scalable,pena2007towards} stated that $CPC(C)$ learnt by GetPCD may be a superset of the true parents and children of $C$ since some non-child descendants of $C$ are added to $CPC(C)$. Thus GetPC was proposed to remove these non-child descendants using the symmetry check.

As for finding the spouses of $C$, for each feature $X\in CPC(C)$ obtained by GetPC, first, PCMB uses GetPC to find the parents and children of $X$ (i.e., $PC(X)$), then for each feature $Y$ in $PC(X)$, if there exists a subset $S$ within $F\setminus\{C,X,Y\}$ ($S$ was identified and stored in the procedure of GetPCD) such that both $C\indep Y|S$ and $C\nindep Y|S\cup \{X\}$ hold, $Y$ is a spouse of $C$ with regard to $X$. The above procedure of finding the spouses of C is summarized in Algorithm 8~\citep{pena2007towards,aliferis2010local1}.
The study in~\citep{pena2007towards,aliferis2010local1} has shown that if the input $CPC(C)$ and the PC learning algorithm used by Algorithm 8 are correct, Algorithm 8 is complete and sound~\citep{aliferis2010local1}.

\begin{algorithm}[t]\label{alg4-6}
\centering
\caption{The Framework of Spouse Learning}
\begin{algorithmic}[1]
\STATE \textbf{Input}: $C$,  $CPC(C)$, and $Sepset (X)$ for each feature $X$ in $F$\\
 \textbf{Output}: Spouses of $C$ ($SP(C)$)
\STATE $SP(C)=\emptyset$;
\FOR {each feature $X$ in $CPC(C)$}
    \STATE Find $CPC(X)$ using a PC learning algorithm (e.g., MMPC)
       \FOR {each feature $Y$ in $CPC(X)\setminus\{C\cup CPC(C)\}$}
           \IF {$Y\nindep C|X\cup Sepset(Y)$}
                \STATE $SP(C)=SP(C)\cup Y$;
\ENDIF
\ENDFOR
\ENDFOR
\STATE Output $SP(C)$
\end{algorithmic}
\end{algorithm}


\textbf{IPCMB.}
The Iterative Parent-Child based search of MB (IPCMB) algorithm~\citep{fu2008fast} is quite similar to PCMB. The key difference between them is that IPCMB employs the RecognizePC algorithm~\citep{li2015practical} to find the PC set of $C$. RecognizePC uses a backward strategy as shown in Algorithm 7. Initially, RecognizePC assumes that all features in $F$ are the candidate PC of $C$, that is, $CPC(C)= F$. To remove false positives from $CPC(C)$,  RecognizePC uses conditional independence tests to check each feature in $CPC(C)$  level by level of the cardinality of the conditioning sets, starting with an empty set.

For spouse discovery,  IPCMB adopts the framework in Algorithm 8. Compared to the divide-and-conquer MB learning algorithms discussed above, an additional improvement is that IPCMB embeds the symmetry check before Step 5 in Algorithm 8. That is, for each feature $X$ in $CPC(C)$,  if $CPC(X)$ obtained at Step 4 does not include $C$, IPCMB does not implement Steps 6 to 8 and  moves to the next feature in $CPC(C)$.

\textbf{MBOR.}
The larger the size of a conditioning set in a conditional independence test,  the less reliable is the independence test.
the MB learning algorithms discussed above, such as IAMB (and its variants), MMMB, HITON-MB, and PCMB, may miss true positives due to the unreliability of the conditional independence tests if the conditioning set is large.
In order to increase the data-efficiency and the robustness of MB learning, MBOR (Markov Boundary search using the OR condition)~\citep{de2008novel} was designed to keep the sizes of conditional sets as small as possible during the search.
MBOR consists of the following three steps.
At Step 1, MBOR discovers a superset of the parents and children of $C$ ($PCS(C)$) and a superset of the spouses of $C$ ($SPS(C)$) by severely restricting
the size of a conditioning set $S$ in the tests to $|S|\leq 1$ and $|S|\leq 2$ respectively.  Thus this reduces the risk of missing features that are weakly associated to $C$ and enhances the reliability of the independence tests.
Then at Step 2, MBOR first uses  the MBtoPC algorithm~\citep{de2008novel} to find $CPC(C)$, then learn parents and children of each feature in $PCS(C)\setminus CPC(C)$, and finally applies the OR condition to retrieve a parent or a child of $C$ (called $X$) that $X\notin CPC(C)$, but $C\in PC(X)$. 
Finally, at Step 3, MBOR identifies the spouses of $C$ using the framework in Algorithm 8.

The first difference between MBOR and the existing MB algorithms is that  MBOR applies the ``OR condition"  to consider two features $X$ and $Y$ as neighbors if $Y\in PC(X)$  OR $X\in PC(Y)$. In contrast, MMMB, HITON-MB, and PCMB employ the ``AND condition", which means that two features $X$ and $Y$ are considered as neighbors if $Y\in PC(X)$ AND $X\in PC(Y)$. The OR condition is less strict than the AND condition and makes it easier for true positives to enter the MB.  The second difference is that MBOR finds a superset of the spouses of $C$ from $F \setminus PC(C)$ at Step 1 instead of the union of parents and children of each feature in $PC(C)$. At Step 2, since  MBtoPC employs the simultaneous MB discovery approach to find $PC(C)$, it still suffers from the problem of data inefficiency.

\textbf{STMB.}
For the divide-and-conquer approach, in the spouse discovery step, identifying parents and children of each feature in $CPC(C)$ is the most computationally expensive due to the exhaustive search for conditioning sets.
To mitigate the computational efficiency problem of identifying spouses,  different from the algorithms described above,  the simultaneous MB (STMB) algorithm~\citep{gao2017efficient} presents two new strategies. First, STMB~\citep{gao2017efficient} identifies the spouses of $C$ from $F\setminus CPC(C)$  instead of the union of parents and children of each feature in $CPC(C)$.  Second, STMB removes false positives from $CPC(C)$ using the candidate spouses selected currently instead of using the symmetric check. These two strategies may make STMB more efficient than MMMB, HITON-MB, and IPCMB in the spouse discovery phase, since it  will be computationally expensive or prohibit to learn the union of parents and children of each feature in $CPC(C)$, especially when the size of $CPC(C)$ is large.

Specifically, STMB includes the following four steps.  At Step 1, STMB finds $CPC(C)$ by using the RecognizePC algorithm.
At Step 2, for each feature $X\in CPC(C)$, STMB identifies the spouses of $C$ with regard to $X$ from $F\setminus CPC(C)$ and removes false positives from $CPC(C)$ using the candidate spouses selected at this step 2 alternatively.
At Step 3, STMB removes false positives in $SP(C)$ by using the $CPC(C)\cup SP(C)$ obtained at Step 2. 
At Step 4, STMB removes false positives from $CPC(C)$ by using $SP(C)$ obtained at Step 3. 
After the four steps, STMB obtains the MB of $C$, i.e., $CPC(C)\cup SP(C)$.

Although STMB improves the computational efficiency of identifying spouses, STMB suffers from the problem of data inefficiency at Steps 3 and 4, since at the two steps, STMB uses an entire set as a conditional set instead of a subset exhaustive search. 

\textbf{CCMB.}
The existing MB learning algorithms mainly focus on how to remove the false positives during the MB search process and then make the false positive rate as low as possible. However, they rarely consider the true positives discarded due to incorrect conditional independence tests, leading to a high false negative rate, especially in the presence of insufficient or noise data samples.

To tackle this issue, Wu et al.~\citep{wu2019ccmb}  presented a new concept of \emph{PCMasking} to describe a type of incorrect conditional independence tests in the MB learning process and theoretically analyzed the mechanism behind this type of tests. In the work, \emph{PCMasking} denotes that  the class variable and its children may be independent of each other conditioning on its parents and vice versa due to incorrect  independence tests. 
Based on the theoretical analysis, the cross-check and complement MB (CCMB) learning algorithm was proposed to repair this type of incorrect CI independence tests for accurate MB learning. Specifically, CCMB first learns the PC set of $C$ using a subroutine called FindPC. FindPC is an improved version of the GetPCD algorithm and aims to effectively identify all possible true parents and children of $C$ except for the PC features discarded by FindPC due to the \emph{PCMasking} phenomenon. Then CCMB recovers the discarded PC features using the OR rule based on FindPC. The spouse learning phase  of CCMB is the same as that of PCMB. The drawback of CCMB is that although it significantly reduces the false negative rate, CCMB achieves a little higher false positive rate than the divide-and-conquer algorithms discussed above due to the OR rule.

\subsubsection{Methods of MB learning with interleaving PC and spouse learning}\label{sec333}

Different from the algorithms described above that learn PC and spouses separately, BAMB~\citep{ling2019BAMB} and EEMB~\citep{ling2020EEMB} implement the PC learning phase and the spouse identifying phase alternatively for the trade-off between data efficiency and time efficiency.

\textbf{BAMB.}
The balanced MB learning (BAMB) algorithm~\citep{ling2019BAMB} does not separate PC learning and spouse identifying into two independent phases. It finds the candidate PC and spouse set of $C$ and removes false positives from the candidate set in one go. Specifically, using the IFBS framework, BAMB integrates PC learning and spouse identifying into one procedure. At each iteration, once a new feature is added to the current $CPC(C)$, BAMB is triggered to find the spouses of $C$ ($SP(C)$) with regard to this feature. Then BAMB first uses the found $SP(C)$ to remove false positives from $CPC(C)$, then employs the updated $CPC(C)$ to prune $SP(C)$ in turn. In this way, during the MB search BAMB can keep both $CPC(C)$ and $SP(C)$ as small as possible for achieving a trade-off between data efficiency and time efficiency. However, in the PC  learning and spouse identifying phase, due to false PC's inclusion, many false spouses may enter $SP(C)$, leading to a large size of $SP(C)$. BAMB will perform an subset search in the union of current $SP(C)$ and $CPC(C)$ to remove false PC and spouses respectively, 
 and thus the large size of $\{SP(C)\cup CPC(C)\}$ will make BAMB both time and data inefficient. 

\textbf{EEMB.}
To tackle the drawback of BAMB, the EEMB (efficient and effective MB) algorithm~\citep{ling2020EEMB} breaks BAMB into two independent subroutines: ADDTrue and RMFalse. EEMB first uses the ADDTrue subroutine to learn the candidate PC set and the spouse set, then employs the  RMFalse subroutine for pruning the two sets. In the ADDTrue subroutine, before a candidate PC feature $X$ is added to the current $CPC(C)$, EEMB will test whether $X$ is independent of $C$ using the current $CPC(C)$.  If so, $X$ will be discarded and consider the next candidate PC feature.  If not, EEMB is triggered to identify the spouses of $C$ with regard to $X$ without performing an subset search in the current $SP(C)$.  After this pruning, EEMB will greatly prune the false PC features before the spouse learning phase is triggered and make both $CPC(C)$ and $SP(C)$ keep as small as possible before the RMFalse subroutine runs. In the RMFalse subroutine, EEMB first uses the union of $SP(C)$ and current $CPC(C)$ to prune $CPC(C)$, then removes false positives from $SP(C)$ using the union of the updated $CPC(C)$  and current $SP(C)$.

\subsubsection{Methods of MB learning with relaxed assumptions}\label{sec34}

In this subsection, we will discuss six representative MB learning algorithms for tackling the situation where the faithfulness or causal sufficiency assumption is violated, i.e., KIAMB~\citep{pena2007towards}, TIE*~\citep{statnikov2013algorithms}, SGAI~\citep{yu2017markov},  LCMB~\citep{liu2016swamping}, WLCMB~\citep{liu2016swamping}, and M3B~\citep{yu2018mining}.

\textbf{KIAMB.}
Let $S_1$, $S_2$, $Z$ and $W$ denote four mutually disjoint feature subsets, the composition property assumes that if $S_1\indep S_2|Z$ and $S_1\indep W|Z$ hold, then $X\indep (S_2\cup W)|Z$ holds~\citep{pearl2014probabilistic}. The composition property assumption is much weaker than the faithfulness assumption.
The KIAMB algorithm~\citep{pena2007towards} aims to tackle MB learning when the faithfulness assumption is violated.
The difference between KIAMB and IAMB is that KIAMB allows the user to specify the trade-off between greediness and randomness in the MB search through a randomization parameter $K\in [0,1]$. IAMB greedily adds to $CMB(C)$ the feature with the highest association with $C$ among all features excluding features currently in $CMB(C)$, while KIAMB adds to $CMB(C)$ the features with the highest associations with $C$ in the CanMB set which is a random subset of $CMB(C)$ with size $max(1, \llcorner(|CMB(C)|\cdot K)\lrcorner)$.
$K$ specifies the trade-off between greediness and randomness in the MB search: if setting $K=1$,  KIAMB is reduced to IAMB, while if taking $K=0$, KIAMB is a completely random approach which is expected to identify all the MBs of $C$ with a nonzero probability if running repeatedly for enough number of times.
IAMB and KIAMB are both correct under the composition assumption~\citep{pena2007towards}. However, KIAMB does not guarantee finding all MBs of the class variable under the composition assumption and is computationally more expensive than IAMB because it has to be run multiple times.


\textbf{TIE*.}
Statnikov et al.~\citep{statnikov2013algorithms} relaxed the composition assumption to the local composition assumption and proposed a family of the TIE* (Target Information Equivalence) algorithm for multiple MB learning.
Specifically, the TIE* algorithm mainly includes three steps.  In Step 1, TIE* uses an existing single MB learning algorithm  to learn a $MB(C)$  from a data set $D$ defined on $F$ (i.e., the original distribution) and outputs $MB(C)$. In Step 2, TIE* uses a procedure to generate a new data set $D_{new}$ (i.e., the embedded distribution that is obtained by removing subsets of features of $MB(C)$ from the original distribution $D$). The motivation is that $D_{new}$ may lead to identifying of a new $MB(C)$ that was previously ``invisible" to a single MB learning algorithm since it was ``masked" by another MB of $C$. Next, in Step 3 the MB learning algorithm employed in Step 1 is applied to $D_{new}$, resulting in a new candidate MB of $C$, called $CMB_{new}(C)$ in the embedded distribution. If $CMB_{new}(C)$  is also a MB of $C$ in the original distribution according to a criterion (independence tests or classification accuracy), then $CMB_{new}(C)$ is considered as a new MB of $C$.  Steps 1-3 are repeated until all possible data sets $D_{new}$ generated by the procedure used in Step 2 have been considered.
It has been proved that the TIE* algorithm can output all possible MBs of the class variable in data set when the faithfulness assumption  is violated.

\textbf{SGAI.}
When the faithfulness assumption is violated, it  may not be tractable for TIE* to learn all possible MBs for feature selection due to computational complexities. To deal with this problem, the SGAI (Selection via Group Alpha-Investing) algorithm was proposed~\citep{yu2017markov}. Compared to the standard constraint-based MB learning algorithms discussed above, SGAI combines the MB theory with the idea of classical feature selection.
Instead of an exhaustive search over a large number of MBs in a data set, SGAI presents the concept of a  representative set which consists of the features of all possible MBs. Each member in the representative set  is not a single feature, but a feature set (i.e., a group of features). SGAI first uses the existing MB learning algorithms (e.g., HITIOM-MB) to learn the representative sets. Then SGAI presents a group Alpha-investing procedure to select a best subset from representative sets. The group Alpha-investing procedure is motivated by the Alpha-investing feature selection method~\citep{zhou2006streamwise} and can simultaneously optimize selections within each representative set as well as between those sets to achieve a feature subset that maximizes the predictive power for classification.

Compared to TIE*, SGAI does not  learn all possible MBs from a data set, but chooses a feature subset that maximizes the prediction power for classification instead.
However, when both the numbers of groups in the representative set and features in each group become large, SGAI may not be efficient and effective. Furthermore, since the number of MBs in a data set is not known,  the representative set cannot guarantee to include the features of all possible MBs in the data set.  In this case, the final output of SGAI is not optimal for feature selection.

\textbf{LCMB and WLCMB.}
To tackle incorrect independent tests, in~\citep{liu2016swamping}, the problem of incorrect independent tests is described as swamping and masking. Swamping means a true positive becomes a false negative, while masking means a true negative becomes a false positive.  Based on the KIAMB algorithm, the LRH algorithm~\citep{liu2016swamping} was proposed to tackle the problem of swamping and masking and it is correct under the local composition assumption.

Compared to KIAMB, the innovation of LRH is that  a selection-exclusion-inclusion (SEI) procedure was proposed to search for a candidate MB set of $C$ which contains as few false positives as possible. Specifically, in the SEI procedure, the selection phase selects the candidate MB features of $C$ conditioning on the MB currently selected, then for each feature in this MB set,  the exclusion phase removes this feature if it is independent of $C$ conditioning on its neighbors in the MB set; finally, the inclusion phase chooses the $K$ features in the current MB with the high associations with $C$ as the output of the SEI procedure at each iteration.
Since IAMB and KIAMB remain correct under the local composition assumption, in~\citep{liu2016swamping}, IAMB, KIAMB and LRH were integrated into a framework called LCMB (Local Composition MB).
Furthermore, to tackle the violation of the faithfulness assumption, based on the LCMB framework, WLCMB (Weak Local Composition MB) was proposed~\citep{liu2016swamping}. WLCMB interleaves LCMB with a search-resuming procedure and has a higher computational complexity than LCMB.

\textbf{M3B.}
The M3B (Mining Maximal ancestral graph MB) algorithm was proposed to tackle MB learning using independence tests when the causal sufficiency assumption is violated. A Maximal ancestral graph (MAG) model has been developed to deal with latent common causes without pre-determining the number of latent common causes and their exact locations with respect to other features~\citep{richardson2002ancestral,silva2009hidden,borboudakis2016towards}.
Thus, instead of using DAGs, in~\citep{yu2018mining}, authors adopted the MAG model to represent latent common causes and the concept of MBs. Specifically, the work in~\citep{yu2018mining} first defines the concept of MB of the class variable in a MAG, i.e., MAG MB (MMB), and presents  a theoretical analysis of its properties. Then the M3B algorithm was proposed to learn the MMB of the class variable and it was the first constraint-based algorithm that  was specially designed for MB learning when the causal sufficiency assumption is violated. The M3B algorithm mainly includes two novel methods  to find the MMB of the class variable, the AdjV (Adjacent feature) algorithm using a backward strategy as shown in Algorithm 7 to find the parents and children of the class variable and the RecSearch (Recursive Search) algorithm to discover the remaining features of the MMB of the class variable.

\subsubsection{Methods of MB learning with special purpose}\label{sec335}

In the section, we will discuss the six representative MB learning algorithms for some special purposes, i.e.,  MIMB for identifying a MB of a class variable from multiple data sets~\citep{yu2018discovering}, MCFS for stable predictions with distribution shift~\citep{yu2019MCFS}, MIAMB and MKIAMB for learning a MB of multiple class variables~\citep{liu2018markov}, and BASSUM and Semi-IAMB for weak supervision learning~\citep{cai2011bassum,sechidis2018simple}.

\textbf{MIMB.}
The MB learning algorithms discussed above all  learn MBs from a single observational data set. There has been an increasing availability of interventional data collected from various sources, such as gene knockdown experiments by different labs for studying
the same diseases.
 Recently, Yu et al.~\citep{yu2018discovering} studied the problems of MB learning in multiple interventional (experimental) data sets.
This is the first work systematically studying the conditions for finding the correct MB of a class variable and the conditions for identifying the parents of class variable through MB learning.
Based on the theoretical analysis, authors designed the MIMB (Multiple Interventional  MB) algorithm to learn MB in multiple Interventional data sets.  MIMB  also adopts a divide-and-conquer approach which consists of two new subroutines. One subroutine, called MIPC, was designed for discovering $PC(T)$ from multiple interventional data sets using the IFBS framework as presented in Algorithm 6,  and the other was proposed to identify spouses of $C$ from multiple interventional data sets based on the framework as shown in Algorithm 7 .


\textbf{MCFS.}
To achieve stable predictions for multiple data sets with different distributions,  based on the theoretical results in~\citep{yu2018discovering}, the MCFS (multi-source causal feature selection) algorithm was  proposed~\citep{yu2019MCFS}.  By utilizing the concept of causal invariance~\citep{pearl2009causality,peters2016causal} and mutual information, MCFS formulates the problem of stable predictions in multiple data sets as a search for an invariant set across different data sets. To speed up the search, this work analyzed the upper and lower bounds of the invariant set and made MCFS learn the best invariant set within the bounds for stable predictions. MCFS outperforms some well-known existing feature selection algorithms designed for a single data set. Furthermore, this work demonstrated that for multiple data sets with different distributions, the set of parents of a class variable is the minimal and promising invariant set for stable predictions, while the MB or PC of the class feature may not.

\textbf{MIAMB and MKIAMB.}
 The algorithms described above all focus on learning a MB of a single class variable, e.g.,  $MB(C)$, the MB of $C$. Recently, the work in~\citep{liu2018markov} explored the problem of learning a MB of multiple class variables, e.g., one MB, $MB(C_1, C_2)$ for both class variables $C_1$ and $C_2$. This work first proved that  under the local intersection assumption a MB of multiple class variables can be constructed by simply taking the union of the MBs of the individual class variable excluding the class variables from the union (if they are included in the union).
Then the MB learning problem for multiple class variables was transformed to  a number of MB learning problems of a single class variable. By considering the violation of faithfulness assumption, MIAMB and MKIAMB were  proposed in the work~\citep{liu2018markov}. For a set of class variables of interest, given an ordering which determines which class variable's MB needs to be learned in the current step,  MIAMB and MKIAMB first find a MB of two class variables, and then learns an MB of three class variables and so on until all the class variables are considered.

\textbf{BASSUM.}
In many real-world applications, labelled examples are often expensive to acquire while it is easy to collected unlabelled data examples. To leverage both unlabelled and labelled data to help MB learning (i.e., weak-supervision MB learning), Cai et al.~\citep{cai2011bassum} proposed a novel BAyesian Semi-SUpervised Method (BASSUM). To our best knowledge, BASSUM was the first weak-supervision MB learning algorithm. In the first phase, BASSUM learns the parents and children  and then the spouses of $C$ by taking into account both labelled and unlabelled data examples using a modified version of the $G^2$ test. The modified version of the $G^2$ test can use unlabelled data examples to enhance the reliability of the conditional independence tests. In the second phase, to prune the MB obtained in the first phase using unlabelled data examples,  a concept of \emph{effective feature sets} was proposed. It is a subset of the PC set of $C$ obtained in the first phase. Using the effective feature sets, BASSUM prunes the  PC set of $C$ without accessing the information of $C$ in labeled data examples. 
However,  one weakness of BASSUM is that there are no guarantees that the modified $G^2$ test will follow a chi-squared distribution. BASSUM also cannot deal with the situation of partially labelled samples, i.e., we have a small number of binary labelled data and a vast amount of  unlabelled examples.

\textbf{Semi-IAMB.}
In order to deal with partially labelled data samples, the work in~\citep{sechidis2018simple,sechidis2015markov} first proposed a generalization of the conditional independence tests for partially labelled samples and then extended the work to semi-supervised data which contains a small number of binary labelled data and a large number of unlabelled examples.
In the work, authors proposed some theoretical results of hypothesis testing (e.g., the $G^2$ test) and feature ranking in the partially labelled data environment.

Specially, by assuming that all missing labels are negative,or assuming that they are positive, authors present a surrogate class variable for semi-supervised hypothesis testing. 
That is, let $C_0$ represent assigning 0 to all missing class labels and $C_1$ represent assigning 1 to all missing class labels, authors propose to use surrogate test $X\indep C_0$ or $X\indep C_1$ to replace the true unlabelled class variable test $X\indep C$).
And they have proved that (1) both surrogate tests (i.e., $X\indep C_0$ or $X\indep C_1$) have exactly the same false positive rate  as the ideal test (i.e., $X\indep C$); (2) both surrogate tests will have a higher  false negative rate than the ideal test.
To reduce false negative rate, authors suggested use more data samples (if possible) or prior knowledge of the class probability to determine which one of the two surrogates will have the lower false negative rate.
Moreover,  in the work, it has been proved that both surrogate tests produce exactly the same feature ranking as  $X\indep C$. 
Then based on these theoretical results authors developed the Semi-IAMB algorithm~\citep{sechidis2018simple} which uses the surrogate tests.
However, the theoretical results in the work now only can deal with binary class variables and consequently Semi-IAMB cannot learn the MB of a class variable with more than two classes. In addition, to reduce false negative rate  and improve feature ranking quality, Semi-IAMB requires more data samples and  prior knowledge for surrogate tests.

\begin{table}[t]
\centering
\caption{Representative score-based MB learning algorithms using score criteria}
\begin{tabular}{|l|l|}
\hline
Category                                                                                                                     & Algorithm                  \\ \hline
\multirow{3}{*}{\begin{tabular}[c]{@{}l@{}}Divide-and conquer MB learning\\ (learning PC and spouses separately\\ using a BN structure learning algorithm)\end{tabular}}
                                                                                                                             &SLL~\citep{niinimki2012local}    \\ \cline{2-2}
                                                                                                                             & $S^2TMB$~\citep{gao2017efficient2}    \\ \cline{2-2}
                                                                                                                             & $S^2TMB^+$~\citep{gao2017efficient2}     \\ \hline
\multirow{2}{*}{\begin{tabular}[c]{@{}l@{}}Simultaneous MB learning (learning PC\\ and spouses simultaneously)\end{tabular}}
                                                                                                                             & DMB~\citep{acid2013score}        \\ \cline{2-2}
                                                                                                                             & RPDMB~\citep{acid2013score}        \\ \hline
\multirow{2}{*}{MB learning with relaxed assumptions}
                                                                                                                            & BSS-MB~\citep{masegosa2012bayesian}  \\ \cline{2-2}
                                                                                                                            & LMB-CSEM~\citep{gao2016constrained}   \\ \hline
\end{tabular}
\label{tb4-1}
\end{table}

\section{Score-based methods}\label{sec4}

This type of methods employs score-based BN structure learning algorithms  to learn the MB or PC of the class variable instead of using independence tests. Table~\ref{tb4-1} summarizes the representative score-based MB learning algorithms. Score-based MB learning algorithms are not  the focus in MB learning research, thus the number of algorithms is much smaller than constraint-based algorithms.

In the following, Section~\ref{sec41} presents the basis of score-based methods. Section~\ref{sec42} gives the brief discussions of score-based methods.  Section~\ref{sec43} extensively reviews the representative score-based methods.

\subsection{Basis of score-based methods}\label{sec41}

Given a data set $D$, score-based BN learning algorithms aim to find the structure of the BN, i.e. the DAG, that maximizes a scoring function, which is usually defined as a measure of fitness between the DAG and $D$. They use the scoring function in combination with a greedy search method in order to measure the goodness of each explored structure from the space of feasible solutions.

The representative scoring functions designed based on different principles include K2~\citep{cooper1992bayesian}, BDeu~\citep{buntine1991theory}, BDe~\citep{heckerman1995learning}, BIC/MDL~\citep{lam1994learning,schwarz1978estimating}, AIC~\citep{akaike1974new}, and MIT~\citep{campos2006scoring}.
The score-based BN learning problem can be formulated as: given $D$, learning a DAG $G*$ such that
$G*=\arg\max_{G\subseteq O}f(G:D)$ where $f (G:D)$ is the scoring function and $O$ is the family of all possible DAGs defined on $D$. A desirable property for a scoring function is the decomposability that enables to compute the global score of a DAG by aggregating local scores.  $f (G:D)$ is decomposable if the score assigned to a structure can be expressed as a combination of local scores of each node and its parents in $G$: $f(G:D)=\sum_{V_i\in V}f(V_i; pa_G(V_i): D_{V_i, pa_G(V_i)})$.

Since scoring functions are decomposable, the main idea of score-based MB learning algorithms is to learn a DAG of the features currently selected, $C$, and a new feature, then reads the MB (or PC) from the DAG at each iteration. Thus the score-based algorithms can distinguish parents from children of the class variable during MB learning, while the constraint-based algorithms cannot.

\subsection{Overview of score-based methods}\label{sec42}

Existing score-based MB learning algorithms are mainly the score-based variants of the constraint-based MB learning algorithms. Through learning a DAG around a class variable, these algorithms read the MB of the class variable from the DAG. Since existing score-based MB learning algorithms are motivated from constraint-based methods, in Table~\ref{tb4-1},  we categorize these algorithms into three types: divide-and-conquer MB learning, simultaneous MB learning, and MB learning with relaxed assumptions.

The SLL algorithm~\citep{niinimki2012local}  is a score-based variant of the divide-and-conquer MB learning algorithms. In the PC learning and spouse identifying phases, SLL employs a BN structure learning algorithm to learn PC and spouses separately.  To removing false positives, SLL implements the  symmetric check using the AND rule to remove false positives in the found PC set, while  the  symmetric check using the OR rule to remove false positives in the found spouse set. The symmetric check makes SLL computationally expensive  as the size of the MB of the class variable becomes large.

To improve the search efficiency of SLL, the $S^2$TMB algorithm~\citep{gao2017efficient2} was proposed which is a score-based variant of STMB. $S^2$TMB learns the spouses of $C$ from $F\setminus CPC(C)$ instead of the union of parents and children of each feature in $PC(C)$, and employs the found spouses and PC to remove false positives instead of the symmetric check.
 $S^2$TMB$^+$ is an improved version of $S^2$TMB for further improving the computationally efficiency of $S^2$TMB.

Different from SLL and $S^2$TMB, DMB and RPDMB~\citep{acid2013score} do not divide MB learning into the PC learning step and the spouse identifying step. Instead, DMB and RPDMB learn PC and spouses of $C$ simultaneously. These two algorithms only need to learn a DAG around the class variable to obtain a MB of class feature instead of learning many local DAGs.

When the the faithfulness assumption is violated, the BSS-MB algorithm~\citep{masegosa2012bayesian} was proposed to learn multiple MBs using a score criterion. It is a score-based variant of KIAMB for learning multiple MBs
When the the causal sufficiency assumption is violated, the LMB-CSEM algorithm~\citep{gao2016constrained} was the first score-based algorithm to learn the MB of $C$ with latent features in a DAG.
 Using score-based methods, BSS-MB does not guarantee finding all possible MBs, and it does not show significantly advantages over KIAMB or TIE* in terms of time efficiency and learning accuracy.  LMB-CSEM needs to use the EM algorithm to tackle the missing values of latent variables, and thus it will be computationally expensive when the size of data samples is large.

In summary, so far it is not easy to use score criteria for MB learning when the faithfulness or causal sufficiency  are violated, and these algorithms may suffer the computational problem of BN structure learning and they are still  based on the framework of the constraint-based MB learning. And they algorithms do not show significant advantages over the constraint-based MB learning algorithms, and they have not attached as much attention as constraint-based methods in the MB learning research.

\subsection{Detailed review of score-based methods}\label{sec43}

\subsubsection{Divide-and conquer methods}\label{sec431}

In this subsection,  we discuss the three representative score-based methods with the divide-and conquer  strategy as follows. 

\textbf{SLL.}
The SLL (Score-based Local Learning) algorithm~\citep{niinimki2012local} first learns the PC set of a class variable as shown in Algorithms 9 and 10, and second identifies the spouses of the class variable as shown in Algorithm 11. Specifically, SLL includes the following four steps.

\begin{algorithm}[t]\label{alg5-1}
\centering
\caption{Candidate PC leaning by SLL}
\begin{algorithmic}[1]

\STATE \textbf{Input}: Feature set $F$ and the class variable $C$\\
 \textbf{Output}: $CPC(C)$

\STATE $CPC(C)=\emptyset$;
\REPEAT
\STATE Select a feature $X\in F$;
\STATE $F=F\setminus X$;
\STATE //using an existing score-based BN learning algorithm
\STATE Learning a DAG on the set $CPC(C)\cup \{X\}\cup \{C\}$;
\STATE Obtain $CPC(C)$ from the learnt DAG;
\UNTIL{ $F$ is empty};
\STATE Output $CPC(C)$.
\end{algorithmic}
\end{algorithm}

\begin{algorithm}[t]\label{alg5-2}
\centering
\caption{PC leaning with symmetric checks by SLL }
\begin{algorithmic}[1]

\STATE \textbf{Input}: Feature set $F$ and the class variable $C$\\
 \textbf{Output}: $PC(C)$

\STATE Find $CPC(C)$ using Algorithm 9;
\STATE //Symmetric check whether $C\in PC(X) (X\in CPC(C))$
\REPEAT
\STATE Select a feature $X\in CPC(C)$;
\STATE $CPC(C)=CPC(C)\setminus X$;
\STATE Obtain $CPC(X)$ using Algorithm 9;
\IF {$C\notin CPC(X)$}
\STATE $CPC(C)=CPC(C)\setminus X$;
\ENDIF
\UNTIL {$CPC(C)$ is empty};
\STATE $PC(C)= CPC(C)$;
\STATE Output $PC(C)$.
\end{algorithmic}
\end{algorithm}

\begin{algorithm}[t]\label{alg5-3}
\centering
\caption{Spouse learning  by SLL }
\begin{algorithmic}[1]

\STATE \textbf{Input}: Feature set $F$ and $PC(C)$\\
 \textbf{Output}: $SP(C)$

\STATE Find candidate spouses $CSP(C)$, i.e., PC of each feature in $PC(C)$ using Algorithm 10;
\STATE $CSP(C)=CSP(C)\setminus PC(C)\cup C$; $SP(C)=\emptyset$;
\REPEAT
\STATE Select a feature $X\in CSP(C)$;
\STATE $CSP(C)=CSP(C)\setminus X$;
\STATE Learning a DAG on the set $PC(C)\cup \{X\}cup \{C\}\cup SP(C)$;
\STATE Obtain SP(C) from the learnt DAG;
\UNTIL {$CSP(C)$ is empty};
\STATE Output $SP(C)$.
\end{algorithmic}
\end{algorithm}

\begin{itemize}

\item (Step 1) Finding candidate PC of $C$.  In Algorithm 9, initially $CPC(C)=\emptyset$. At each iteration, SLL randomly selects a feature $X\in F$ and removes $X$ from $F$, then uses a score-based BN learning algorithm, such as those in~\citep{chickering2002optimal,koivisto2004exact,rohekar2018bayesian}, to learn a DAG of the set $(C\cup CPC(C)\cup X)$. SLL obtains a new $CPC(C)$ from the learnt DAG. The final $CPC(C)$ will be obtained until the set $F$ is empty.

\item (Step 2) Symmetry checks for pruning $CPC(C)$.  SLL uses a score-based variant of symmetric checks as shown in Algorithm 10.  SLL learns the PC of each feature $X$ in $CPC(C)$ using Algorithm 9. If $C\notin CPC(X)$, SLL removes $X$ from the $CPC(C)$.

\item (Step 3) Identifying  the spouses of $C$ as shown in Algorithm 11. Let  $SP(C)=\emptyset$.  SLL first uses Algorithm 9 to find the union of PC of each feature in $PC(C)$ obtained in Step 2 as the candidate spouses of $C$, called $CSP(C)\setminus PC(C)\cup C$.  Then for each feature $X$ in this union, SLL learns a DAG of $(C\cup PC(C)\cup X\cup SP(C))$ and obtains a new $SP(C)$ from the learnt DAG until the union is empty.

\item (Step 4) Finalizing spouses of $C$ by the OR-rule symmetry constraint.  In this step, SLL performs symmetric checks for finalizing spouses. That is, if $C\in SP(X)$ but $X\notin SP(C)$, using the OR rule, $X$ should be added to $SP(C)$. SLL first uses Algorithm 11 to find $SP(C)$. Then SLL learns the spouses of all features in $F\setminus PC(C)$ using Algorithm 11. If the spouse set of a feature includes $C$, the feature will be added to $SP(C)$. The symmetric check will be computational expensive when the size of $F\setminus PC(C)$ is large.

\end{itemize}

SLL is computationally expensive to lean DAGs for symmetric checks in Steps 2 and 4, especially with a large size of the MB of $C$.

\textbf{$S^2$TMB.}
The $S^2$TMB (Score-based Simultaneous MB) algorithm aims to improve the search efficiency of SLL by removing the symmetry checks in both PC and spouse search steps (i.e., Steps 2 and 4 of SLL). $S^2$TMB mainly consists of the following two steps.
\begin{itemize}
\item (Step 1) $S^2$TMB shares the same Step 1 as SLL for learning $CPC(C)$.

\item (Step 2) Pruning $CPC(C)$ and identifying $SP(C)$. Let $SP(C)=\emptyset$ and $R=F\setminus CPC(C)$. $S^2$TMB learns the spouses of $C$ (i.e., $SP(C)$) from $R$ instead of the union of parents and children of each feature in $PC(C)$. It prunes $CPC(C)$ and identifies $SP(C)$ simultaneously at Step 2.
For each feature $X\in R$, $S^2$TMB learns iteratively a DAG of the subset of $C\cup CPC(C)\cup SP(C)\cup X$, and prunes $CPC(C)$ and obtains $SP(C)$ using the learnt DAG, until $R$ is empty.
\end{itemize}

\textbf{$S^2$TMB$^+$.}
However, in Step 2, the size of $CPC(C)\cup SP(C)$  may grow uncontrollably large, leading to the same expensive computational cost as BN structure learning.
 To make the size of BN structures learnt at each iteration as small as possible, $S^2$TMB$^+$ decomposes Step 2 of $S^2$TMB into two steps as follows. At Step 2(a), $S^2$TMB$^+$ only learns a DAG of $C\cup CPC(C)\cup X$ to prune $CPC(C)$ and obtain $SP(C)$ instead of $C\cup CPC(C)\cup SP(C)\cup X$. And Step 2(b) uses the features in $SP(C)$ one by one to prune both $CPC(C)$ and $SP(C)$.

\begin{itemize}
\item (Step 1)  $S^2$TMB$^+$ uses the same method as $S^2$TMB for learning $CPC(C)$.

\item (Step 2a) Pruning $CPC(C)$ and learning $SP(C)$.
 Let $SP(C)=\emptyset$ and $R=F\setminus CPC(C)$ initially. For each feature $X\in R$, $S^2$TMB$^+$ learns iteratively a  DAG of the subset of $C\cup CPC(C)\cup X$ instead of $C\cup CPC(C)\cup SP(C)\cup X$ , and prunes $CPC(C)$ and obtains $SP(C)$ using the learnt DAG, until $R$ is empty.


\item (Step 2b) Pruning spouses and $CPC(C)$. In this step,  let $R=SP(C)$ and $SP(C)=\emptyset$. For each feature $X$ in $R$,  $S^2$TMB$^+$ learns iteratively a  DAG of the subset $C\cup CPC(C)\cup X\cup SP(C)$, then obtain $CPC(C)$ and $SP(C)$ from the learnt DAG, until $R$ is empty.

\end{itemize}


\subsubsection{Simultaneous MB learning methods }\label{sec432}

DMB and RPDMB~\citep{acid2013score} are different from SLL and $S^2$TMB. They do not divide MB learning into the PC learning step and the spouse identifying step. Instead, DMB and RPDMB learn PC and spouses of $C$ simultaneously. These two algorithms only need to learn a local DAG around the class variable to obtain a MB of class feature instead of learning many local DAGs.
Specially, they first define two restricted search spaces, that is, CDAGs (Class-focused DAGs, see Definition 1 in~\citep{acid2013score}) and CRPDAGs (Class-focused Restricted Partially Directed Acyclic Graphs, see Proposition 1 in~\citep{acid2013score}).
Then starting from an empty graph, using the hill-climbing-based search operators proposed in~\citep{acid2005learning},  DMB carries out a local search in the space of CDAGs while RPDMB implements the search in the space of CRPDAG. Both algorithms terminate until the scoring function does not improve. Finally, the two algorithms read $MB(C)$ from the obtained graphs respectively.  Compared to  SLL and $S^2$TMB, DMB and RPDMB do not need to learn DAGs many times. But the problem is that how to obtain the two restricted search spaces is not clear in the paper~\citep{acid2005learning}. If the size of the restricted search space is large, the computational cost of learning DAGs  will be expensive.

\subsubsection{MB learning methods  with relaxed assumptions}\label{sec433}

In this section, we will discuss two representative score-based MB learning algorithms, BSS-MB and LMB-CSEM, for tackling the situations when the assumptions of faithfulness or causal sufficiency is violated.

\textbf{BSS-MB.}
The BSS-MB (Bayesian stochastic search of MBs) algorithm~\citep{masegosa2012bayesian} is a score-based variant of KIAMB for learning multiple MBs when the the faithfulness assumption is violated.
BSS-MB adopts a strategy similar to that used by KIAMB with $K=0$, but it uses a Bayesian score framework instead of conditional independence tests.  In the growing phase, BSS-MB incrementally adds new features to the candidate MB sets by computing the posterior probability of a conditional independence statement. In the shrinking phase, BSS-MB removes from the MB sets the false positives identified using a Bayesian score.
In addition, compared to KIAMB, each MB set found by BSS-MB has an associated score which measures how well this feature subset acts as a MB.

\textbf{LMB-CSEM.}
LMB-CSEM~\citep{gao2016constrained} treats identifying the latent features included in the MB as a missing value problem. It first assumes the existences of  latent features in the MB of $C$, then assigns these latent features into different non-overlapping latent subspaces.
Within each subspace, LMB-CSEM employs a constrained structure expectation-maximization (CSEM) algorithm to greedily learn the MB with latent features. Then the final optimal MB is obtained from the optimal MBs within each subspace. Specifically, LMB-CSEM has three major steps. At Step 1, LMB-CSEM uses a standard MB discovery algorithm to find a MB of $C$ from observed features as the baseline. At Step 2, using the baseline MB set,
LMB-CSEM  employs the CSEM algorithm to learn a MB with one latent feature within each subspace. At Step 3,  if the score of the learned MB with one latent feature in one subspace is higher than that of the baseline MB, the learned MB will be considered as a new baseline MB, and Steps 2 and 3 are repeated to learn another latent feature until adding more latent features into the learned MB no longer improves the MB score or violates the size constraint.

\section{Methods for distinguishing parents from children}\label{sec5}

Distinguishing  direct causes and direct effects of a class variable is critical to the prediction of the consequence of the actions/interventions or the construction of a robust prediction model in various big data analytics and machine learning tasks.  Existing studies have illustrated that the direct causes of a class variable is the minimal and stable subset for classification when training data and testing data have different distributions~\citep{yin2008partial,yu2019MCFS}.
However, existing constraint-based MB or PC learning methods cannot distinguish parents from children, as they are based on conditional independence tests only, which does not indicate directions of relationships. 
Table~\ref{tb5-1} summaries three approaches for addressing this problem, global  BN structure learning, local structure learning, and local structure learning with experimental data. Note that in Table~\ref{tb5-1} we only summarize the representative local-to-global BN structure learning algorithms since global BN structure learning often is computationally prohibitive~\citep{ghoshal2017learning}.

\textbf{Global structure learning.}
A global structure learning algorithm first learns a global BN structure, then reads parents and children of $C$ from the learnt BN~\citep{nie2017efficient,benjumeda2019learning,campos2011efficient,yehezkel2009bayesian}. However, global BN structure learning is a NP-hard problem due to the huge candidate structure in the search space~\citep{chickering2004large,ghoshal2017learning}.

Accordingly, local-to-global structure leaning methods were proposed.  This type of algorithms first identifies each feature's MB (or PC) using the existing causality-based feature selection methods, then constructs a DAG skeleton (i.e., a undirected graph)  using the found MBs (or PCs), and finally orients the edges of the skeleton using independence tests or score criteria. Several local-to-global structure learning approaches have been proposed to reduce the complexity of global BN structure learning, such as GSBN (Growing-shrinking BN)~\citep{margaritis2000bayesian}, MMHC (Max-Min Hill Climbing)~\citep{tsamardinos2006max}, and SLL+C/G~\citep{niinimki2012local}.
In order to improve the MB learning efficiency, Pellet et al.~\citep{pellet2008using} employed the Relief feature selection algorithm and proposed the TC (Total Conditioning) algorithm. This algorithm uses the Relief feature selection algorithm to identify an approximate MB, then employs conditional independence tests to orient edges.
 Gao et al.~\citep{gao2017local} recently proposed a novel GGSL (Graph Growing Structure Learning) algorithm for local-to-global structure learning. Instead of finding the MBs of all features first, GGSL starts with a randomly selected feature, and then gradually expands the learned structure through a series of local structure learning steps using score-based MB learning algorithms. 
Later, Gao et al.~\citep{gao2018parallel} proposed a parallel Bayesian network structure learning algorithm, call PSL, using multiple local structure learning agents at the same time.
These studies have shown that the local-to-global BN learning approach can deal with a data set with thousands of features.

\begin{table}[t]
\centering
\caption{Representative methods for distinguishing parents from children}
\begin{tabular}{|l|l|}
\hline
Category                                                                                                                     & Algorithm                  \\ \hline
\multirow{6}{*}{Global structure learning}
                                                                                                                             &GSBN~\citep{margaritis2000bayesian}    \\ \cline{2-2}                                                                                                                                                                                                                                                    
                                                                                                                             &MMHC~\citep{tsamardinos2006max}    \\ \cline{2-2}
                                                                                                                             &TC~\citep{pellet2008using}\\ \cline{2-2}
                                                                                                                             &SLL+C~\citep{niinimki2012local}\\ \cline{2-2}
                                                                                                                             &SLL+G~\citep{niinimki2012local}\\ \cline{2-2}
                                                                                                                             &GGSL~\citep{gao2017local}\\ \cline{2-2}
                                                                                                                             &PSL~\citep{gao2018parallel}\\ \hline                                                                                                                        
                                                                                                                             
\multirow{3}{*}{Local structure learning}
                                                                                                                             &PCD-by-PCD~\citep{yin2008partial}        \\ \cline{2-2}
                                                                                                                             &CMB~\citep{gao2015local}  \\ \cline{2-2}
                                                                                                                             &MB-by-MB~\citep{wang2014discovering}        \\ \hline
\multirow{2}{*}{\begin{tabular}[c]{@{}l@{}}Local structure learning \\with experimental data\end{tabular}}
                                                                                                                            &ODLP~\citep{statnikov2015ultra}  \\ \cline{2-2}
                                                                                                                            &ODLP*~\citep{statnikov2015ultra}   \\ \hline
\end{tabular}
\label{tb5-1}
\end{table}

\textbf{Local structure learning.}
Since in many real-world applications we are only interested in the causal relationships around a class variable (e.g., causal genes of lung cancer in a gene data set), it is not  necessary to waste time and memory to learn a global BN structure.  Then several local learning algorithms have been designed for distinguishing direct causes and direct effects of a class variable, such as the CMB (Causal Markov Blanket)~\citep{gao2015local}, PCD-by-PCD (Parents, Children and some Descendants)~\citep{yin2008partial}, and MB-by-MB algorithms~\citep{wang2014discovering}. Given a class variable, these algorithms first find a MB or PC of the class variable and constructs a local structure among the class variable and the features in the MB or PC, then sequentially finds the MB or PC of the features connected to the class variable and simultaneously constructs local structures along the paths starting from the class variable until the parents and children of the class variable have been distinguished or it is clear that the parents and children cannot be distinguished further by continuing the process.
For a class variable in a large network, the local learning algorithms are able to greatly reduce CPU time through finding only a local structure around the class variable compared with the entire BN network learning approaches. Since all existing local learning methods employ the standard MB or PC learning algorithms, they inherit the drawbacks of these algorithms. Furthermore,  the local learning algorithms need to sequentially find the MBs or PCs of the features until the causes and effects of the class variable have been distinguished, and thus their time complexity may not be controllable.

\textbf{Local structure learning with experimental data.}
The local structure learning methods discussed above are fully dependent on observational data and may not be able to orient all edges with observational data alone~\citep{aliferis2010local2,statnikov2015ultra}. Statnikov et. al.~\citep{statnikov2015ultra}  proposed the ODLP algorithm for distinguishing direct causes and direct effects of the class variable using both observational and experimental data. In this work, two versions of the ODLP algorithm, ODLP* and ODLP, were proposed.
Under the faithfulness assumption, the ODLP* algorithm first uses the HITON-PC or MMPC algorithm to find the set of PC of the target with observational data, then orients edges using experimental data to distinguish direct causes from direct effects of the class variable. The ODLP* algorithm is sound and complete.
In order to increase the learning accuracy, by relaxing the faithfulness assumption, ODLP employs the TIE* algorithm to learn multiple PC sets of  the class variable, then finds the true parents and children of the class variable from the union of the multiple PC sets with experimental data. The rational behind the idea is that a feature may have multiple sets of parents and children when the faithfulness assumption is violated~\citep{statnikov2013algorithms}.

\section{The toolbox}\label{sec6}

There are several open-source toolboxes for Bayeisan  or causal network learning, such as the well-known BNT in MATLAB \footnote{https://github.com/bayesnet/bnt}\citep{murphy2001bayes}, PGM in R\footnote{http://mensxmachina.org/en/software/probabilistic-graphical-model-toolbox/},
 bnlearn in R\footnote{http://www.bnlearn.com/}\citep{scutari2009learning},
tetrad in JAVA\footnote{https://github.com/cmu-phil/tetrad}\citep{scheines1998tetrad},
and pcalg in R\footnote{https://cran.r-project.org/web/packages/pcalg/index.html}\citep{kalisch2012causal}. However, these toolboxes do not focus on causality-based feature selection, but Bayesian or causal network structure learning. 
For example, the bnlearn toolbox contains the several causality-based feature selection algorithms, such as GSMB, IAMB, Inter-IAMB, Fast-IAMB, MMPC, and HITON-PC, but it aims to use these algorithms for implementing algorithms of BN learning, inference, and classification.
The Causal Explorer package~\citep{statnikov2010causal} is a well-known local causal discovery package in MATLAB, including several representative causality-based feature selection algorithms, but it is not provided with source code and not available for public use now. 

In this paper,  we have developed the CausalFS toolbox for causality-based feature selection. The CausalFS toolbox provides the first comprehensive open-source library for use in C/C++ that implements the state-of-the-art algorithms of causality-based feature selection.
The toolbox is designed to facilitate the development of new algorithms in this exciting research direction and make  it easy to compare new methods and existing ones.
The CausalFS toolbox is available from https://github.com/kuiy/CausalFS.

\begin{figure}[t]
\centering
\includegraphics[height=4.2in,width=4.5in]{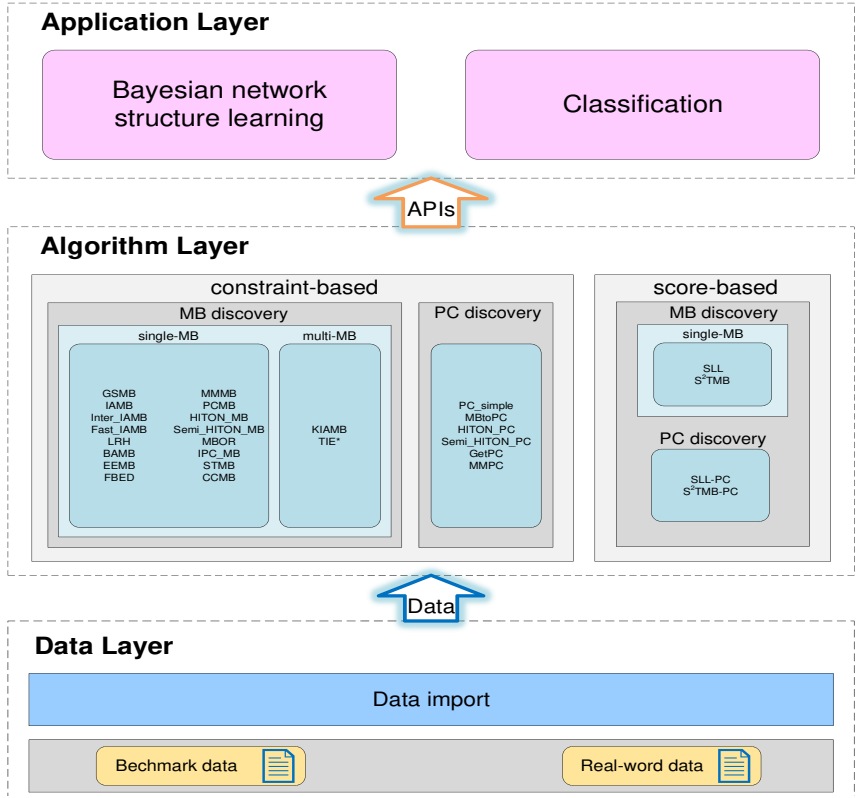}
\caption{Architecture of the causality-based feature selection toolbox}
\label{fig6-1}
\end{figure}

CausalFS was developed in Linux systems. 
The architecture of the CausalFS toolbox in Figure~\ref{fig6-1} contains three layers: application, algorithm, and  data.
The three layers are designed independently. This makes it easy to implement and extend CausalFS. One can easily add a new algorithm to the CausalFS toolbox and share it through the CausalFS framework without modifying the other layers.
In the algorithm layer,  CausalFS mainly implements 28 representative causality-based feature selection methods, including  24 constraint-based algorithms (i.e., 16  algorithms for learning a single MB, 2 algorithms for learning multiple MBs, and 6 algorithms for learning PC and 4  score-based MB and PC learning algorithms.

The algorithm layer of CausalFS can also support local-to-global BN structure learning.
By applying the MB and PC learning algorithms in the algorithm layer, using CausalFS, it is easy to design different  local-to-global structure learning methods. For example, using the MMMB algorithm, we can generate  MMMB based local-to-global structure learning algorithm. All implementation details are included in the detailed documentation available at https://github.com/kuiy/CausalFS, where all algorithms and related data structures are explained in detail.




\section{Evaluations of causality-based feature selection methods}\label{sec7}

In this section, we systematically evaluate the causality-based feature selection algorithms using the CausalFS package in two aspects as follows with synthetic and real-world data sets.  
\begin{itemize}
\item For a synthetic data set, we can read the MB or PC or parents of a feature in the corresponding benchmark BN. We evaluate the quality of the MB or PC of a variable learnt by an algorithm by comparing the MB or PC of the variable with the true MB or PC of the variable in the BN. 


\item For a real-world data set, we evaluate a causality-based feature selection algorithm based on the classification performance of the selected features.
\end{itemize}


All experiments were conducted on a computer with Intel(R) i5-2600, 3.4GHz CPU, and 8GB memory.
In the following experiments, the $G^2$ test is used for independence tests and the significance level for the tests is set to 0.05 and 0.01, respectively. In all tables reporting experiment results, ``-'' denotes that  an algorithm cannot deal with a data set due to expensive computations (the running time of the algorithm exceeded the three-day time threshold and it was stopped).

\subsection{Experiments on synthetic data}\label{sec71}

In this section, we systematically evaluate the algorithms in the CausalFS package using the four standard benchmark BNs as shown in Table~\ref{tab7-1}. 
Using the four benchmark networks, we randomly generate two groups of data, one  including 5 data sets with 500 data instances each, and the other also containing 5 data sets with 5,000 data instances each. Using the two groups of data sets, we evaluate the causality-based feature selection algorithms for learning MB and PC, respectively. The following metrics are used to compare the MB or PC of a variable learnt by a causality-based feature selection algorithms with the true MB or PC of the variable in the BN.

\begin{table}[t]
\centering
\caption{Summary of benchmark BNs}{
\begin{tabular}{cccccccc}
\hline

            & Num. & Num.  & Max In/out- & Min/Max     & Variable \\
Network     & Vars & Edges & Degree      & $|$PCset$|$ &Domain  \\
\hline
Child       & 20   & 25    & 2/7         & 1/8         & 2-6    \\
Alarm       & 37   & 46    & 4/5         & 1/6         & 2-4    \\
Pigs        & 441  & 592   & 2/39        & 1/41        & 3-3    \\
Gene        & 801  & 972   & 4/10        & 0/11        & 3-5    \\
\hline
\end{tabular}}
\label{tab7-1}
\end{table}

\begin{itemize}
\item Precision. The number of true positives in the output (i.e., the features in the output belonging to the true MB or PC of a variable) divided by the number of features in the MB or PC output by an algorithm. 
Precision reports the false positive rate in the output of an algorithm.

\item Recall. The number of true positives in the output divided by the total number of true positives (the size of the true MB or PC) 
Recall reports the true positive rate in the output of an algorithm.

\item $F1$ =$2*precision*recall/(precision+recall)$. It is the harmonic average of the precision and recall, where $F1=1$ is the best case (perfect precision and recall) while $F1=0$ is the worst case.

\item Efficiency. We use the time consumed in seconds to measure the efficiency of an algorithm.


\end{itemize}

For an algorithm, we first use it to learn the MB or PC of each variable in a data set, then compute the average results of recall, precision, F1, and running time over all variables in the data set, and finally we report these average results over five data sets. In all tables of experiment results, a value pair $A/B$ indicates the precision, recall or $F1$ of an algorithm when the significance level of independence tests is set to 0.01 (represented by $A$) and 0.05 (represented by $B$) respectively. 


\begin{table}[!htbp]
\centering
\caption{Results of PC learning methods on synthetic data sets (size=500)\label{tab:four}}{
\tiny
\begin{tabular}{cccccccc}
\\hline

Network                & Algorithm     & F1                 & Precision          & Recall             & Time               \\ \hline
\multirow{8}{*}{Child} & PC-simple     & 0.89/0.89          & 0.94/0.91          & 0.86/0.88          & 0/0                \\
                       & MMPC          & 0.90/0.87           & 0.93/0.87          & 0.89/0.89          & 0/0                \\
                       & HITON-PC      & 0.91/0.88          & 0.94/0.89          & 0.89/0.91          & 0/0                \\
                       & Semi-HITON-PC & 0.91/0.89          & 0.94/0.90           & 0.89/0.91          & 0/0                \\
                       & GetPC         & 0.84/0.84          & 0.93/0.91          & 0.80/0.81           & 0/0                \\
                       & MBtoPC        & 0.86/0.83          & 0.95/0.90           & 0.81/0.80           & 0/0                \\ \cline{2-6}
                       & SLL-PC        & 0.87               & 0.95               & 0.83               & 0.47               \\
                       & S$^{2}$TMB-PC & 0.86               & 0.91               & 0.85               & 0.15               \\ \hline
\multirow{8}{*}{Alarm} & PC-simple     & 0.82/0.84          & 0.89/0.88          & 0.79/0.84          & 0/0                \\
                       & MMPC          & 0.85/0.84          & 0.90/0.84           & 0.83/0.87          & 0/0                \\
                       & HITON-PC      & 0.84/0.83          & 0.89/0.83          & 0.83/0.87          & 0/0                \\
                       & Semi-HITON-PC & 0.85/0.84          & 0.89/0.85          & 0.83/0.87          & 0/0                \\
                       & GetPC         & 0.77/0.82          & 0.88/0.89          & 0.73/0.80           & 0/0                \\
                       & MBtoPC        & 0.86/0.85          & 0.95/0.92          & 0.81/0.82          & 0/0                \\ \cline{2-6}
                       & SLL-PC        & 0.91               & 0.93               & 0.91               & 0.38               \\
                       & S$^{2}$TMB-PC &  0.88              & 0.87               & 0.93               & 0.31               \\ \hline
\multirow{8}{*}{Pigs}  & PC-simple     & 0.99/0.95          & 0.98/0.93          & 1/1                & 0.02/0.03          \\
                       & MMPC          & 0.91/0.77          & 0.86/0.66          & 1/1                & 0/0.01             \\
                       & HITON-PC      & 0.91/0.77          & 0.86/0.66          & 1/1                & 0.01/0.01          \\
                       & Semi-HITON-PC & 0.91/0.77          & 0.87/0.67          & 1/1                & 0.01/0.01          \\
                       & GetPC         & 0.99/0.95          & 0.98/0.92          & 1/1                & 0.04/0.05          \\
                       & MBtoPC        & 0.93/0.89          & 0.97/0.90           & 0.92/0.92          & 0.02/0.03          \\ \cline{2-6}
                       & SLL-PC        & -                  & -                  & -                  & -                  \\
                       & S$^{2}$TMB-PC & -                  & -                  & -                  & -                  \\ \hline
\multirow{8}{*}{Gene}  & PC-simple     & 0.96/0.91          & 0.97/0.89          & 0.96/0.96          & 0.01/0.02          \\
                       & MMPC          & 0.83/0.70           & 0.79/0.60          & 0.92/0.93          & 0.01/0.01          \\
                       & HITON-PC      & 0.83/0.70           & 0.79/0.60          & 0.92/0.93          & 0.01/0.01          \\
                       & Semi-HITON-PC & 0.83/0.71          & 0.79/0.61          & 0.92/0.93          & 0.01/0.01          \\
                       & GetPC         & 0.96/0.94          & 0.97/0.94          & 0.95/0.96          & 0.03/0.04          \\
                       & MBtoPC        & 0.86/0.80           & 0.91/0.83          & 0.84/0.85          & 0.03/0.05          \\ \cline{2-6}
                       & SLL-PC        & -                  & -                  & -                  & -                  \\
                       & S$^{2}$TMB-PC & -                  & -                  & -                  & -                  \\

\hline
\end{tabular}}
\label{tab7-2}
\end{table}

\begin{table}[!htbp]
\centering
\caption{Results of PC learning methods on synthetic data sets (size=5,000)\label{tab:four}}{
\tiny
\begin{tabular}{cccccccc}
\hline

Network                & Algorithm     & F1                 & Precision          & Recall             & Time               \\ \hline
\multirow{8}{*}{Child} & PC-simple     & 1/0.96             & 0.99/0.94          & 1/1                & 0.01/0.02          \\
                       & MMPC          & 0.98/0.92          & 0.96/0.87          & 1/1                & 0.01/0.01          \\
                       & HITON-PC      & 0.98/0.92          & 0.97/0.88          & 1/1                & 0.01/0.01          \\
                       & Semi-HITON-PC & 0.98/0.92          & 0.97/0.88          & 1/1                & 0.01/0.01          \\
                       & GetPC         & 1/1                & 1/0.99             & 1/1                & 0.05/0.05          \\
                       & MBtoPC        & 0.97/0.93          & 0.98/0.92          & 0.97/0.97          & 0.02/0.02          \\ \cline{2-6}
                       & SLL-PC        & 0.97               & 1                  & 0.94               & 5.2                \\
                       & S$^{2}$TMB-PC & 0.98               & 0.99               & 0.97               & 1.51               \\ \hline
\multirow{8}{*}{Alarm} & PC-simple     & 0.98/0.97          & 0.97/0.96          & 0.99/0.99          & 0.02/0.02          \\
                       & MMPC          & 0.97/0.93          & 0.97/0.90           & 0.99/0.99          & 0.01/0.01          \\
                       & HITON-PC      & 0.97/0.92          & 0.97/0.89          & 0.99/0.99          & 0.01/0.01          \\
                       & Semi-HITON-PC & 0.97/0.93          & 0.97/0.90           & 0.99/0.99          & 0.01/0.01          \\
                       & GetPC         & 0.98/0.98          & 1/1                & 0.97/0.97          & 0.04/0.04          \\
                       & MBtoPC        & 0.98/0.96          & 0.99/0.97          & 0.97/0.97          & 0.03/0.03          \\ \cline{2-6}
                       & SLL-PC        & 0.97               & 0.96               & 0.98               & 3.49               \\
                       & S$^{2}$TMB-PC & 0.95               & 0.94               & 0.98               & 2.83               \\ \hline
\multirow{8}{*}{Pigs}  & PC-simple     & 0.99/0.95          & 0.98/0.93          & 1/1                & 0.64/0.78          \\
                       & MMPC          & 0.91/0.78          & 0.86/0.68          & 1/1                & 0.19/0.2           \\
                       & HITON-PC      & 0.91/0.78          & 0.86/0.68          & 1/1                & 0.24/0.26          \\
                       & Semi-HITON-PC & 0.91/0.79          & 0.86/0.69          & 1/1                & 0.29/0.3           \\
                       & GetPC         & 0.99/0.97          & 0.99/0.96          & 1/1                & 10.23/10.43        \\
                       & MBtoPC        & 0.97/0.95          & 0.98/0.94          & 0.99/0.99          & 0.52/0.76          \\ \cline{2-6}
                       & SLL-PC        & -                  & -                  & -                  & -                  \\
                       & S$^{2}$TMB-PC & -                  & -                  & -                  & -                  \\ \hline
\multirow{8}{*}{Gene}  & PC-simple     & 0.98/0.92          & 0.97/0.90           & 0.99/0.98          & 0.40/0.59           \\
                       & MMPC          & 0.83/0.72          & 0.78/0.62          & 0.94/0.94          & 0.12/0.13          \\
                       & HITON-PC      & 0.83/0.72          & 0.78/0.62          & 0.94/0.94          & 0.15/0.19          \\
                       & Semi-HITON-PC & 0.83/0.72          & 0.78/0.63          & 0.94/0.94          & 0.15/0.19          \\
                       & GetPC         & 0.98/0.96          & 0.98/0.95          & 0.99/0.98          & 0.64/0.87          \\
                       & MBtoPC        & 0.91/0.90           & 0.91/0.89          & 0.93/0.93          & 0.69/1.22          \\ \cline{2-6}
                       & SLL-PC        & -                  & -                  & -                  & -                  \\
                       & S$^{2}$TMB-PC & -                  & -                  & -                  & -                  \\
 \hline
\end{tabular}}
 \label{tab7-3}
\end{table}

\textbf{Results of PC learning.}
From Tables~\ref{tab7-2} and~\ref{tab7-3},  for the time efficiency, we have the following observations.

\begin{itemize}
\item An algorithm using a forward strategy for PC learning may be faster than an algorithm using the backward strategy.
Tables~\ref{tab7-2} and~\ref{tab7-3} illustrate that among the six constraint-based PC learning methods, MMPC, HITON-PC, and semi-HITON-PC are very competitive in terms of efficiency, but they are faster than the other PC algorithms. The explanation is that MMPC, HITON-PC, and semi-HITON-PC use a forward strategy, while Recongize-PC uses a backward strategy. 
\item The symmetry check makes a PC learning algorithm computationally expensive. GetPC uses a forward strategy, but it is not efficient since it employs the symmetry check  using the AND rule in the PC learning phase. MBtoPC is slower than the other algorithms, because it performs the symmetry check using the OR rule. SLL-PC implements the symmetry check for both PC and spouse learning. S$^{2}$TMB-PC is faster than SLL-PC since it does not perform the symmetry check.
\item A score-based PC learning algorithm may be slower than a constraint-based method due to the computational  complexity of the BN structure learning. 
In Tables~\ref{tab7-2} and~\ref{tab7-3}, both score-based methods are slower than any constraint-based PC learning algorithms and failed to produce results for the two large-sized networks before they were terminated after three days.
\end{itemize}

In terms of learning performance, the following conclusions are obtained.
\begin{itemize}
\item  A PC learning algorithm using the backward strategy or the symmetry check can remove more false positives, especially on large-sized BN networks.
On the two small-sized networks, Recongize-PC, GetPC, MBtoPC are very competitive with MMPC, HITON-PC, and semi-HITON-PC in terms of precision and recall. On the two large-sized networks, Recongize-PC, GetPC, and MBtoPC significantly outperform MMPC, HITON-PC, and semi-HITON-PC in terms of precision.
This indicates that on the large-sized networks, learning PC using a backward strategy or the symmetry check can remove much more false positives. 

\item A PC learning algorithm using the backward strategy or the symmetry check is not affected by the significance level in terms of precision (i.e., false positive rate).
In Tables~\ref{tab7-2} and~\ref{tab7-3}, the recall metric is not affected much by the significance level parameter, but the precision metrics of  MMPC, HITON-PC, and semi-HITON-PC are influenced greatly by this parameter, while and Recongize-PC, GetPC, and MBtoPC are not affected.
Meanwhile, for the learning performance, SLL-PC (performing the symmetry check) and S$^{2}$TMB-PC are very competitive to PC-simple, GetPC, and MBtoPC on the two small-sized networks.
\end{itemize}

In summary,  the backward strategy or the symmetry check is a double-edged sword. They will make a PC learning algorithm computationally expensive, but they are able to make the PC learning algorithms output a more accurate MB or PC set.


\textbf{Results of MB learning.}
From Tables~\ref{tab7-4} and~\ref{tab7-5}, we have the following conclusions.

\begin{itemize}
\item Existing constraint-based simultaneous MB learning algorithms are inferior the other MB learning algorithms.
From Tables~\ref{tab7-4} and~\ref{tab7-5}, we can see that regardless of sample size (500 or 5000), on the four BN networks, the six simultaneous MB learning algorithms are inferior to the remaining ten MB learning algorithms, especially on the two large-sized BN networks. Meanwhile, GSMB has the worst precisions and recalls among all rivals.
STMB has the lowest precision among all the MB learning algorithms except for GSMB. The explanation is that STMB adopts the same strategy as the simultaneous MB discovery algorithms in the false positive removal phase and this leads to a large data sample requirements.

\item The backward strategy and the symmetry check  make a MB learning algorithm more accurate using a large-sized data samples. 
Since GetPC, Recongize-PC, and MBtoPC outperform the other PC learning algorithms as shown in   Tables~\ref{tab7-3} and~\ref{tab7-4}, their corresponding MB learning algorithms,  PCMB, IPCMB, and MBOR, are better than the other constraint-based 11 MB learning algorithms on the two large-sized BN networks.
 For example,  IPCMB achieves the best performance on the two large networks, since IPCMB not only adopts the Recongize-PC algorithm to learn the PC of each variable, but also it  performs the symmetry check in the spouse discovery procedure (Recongize-PC gets the best performance as shown in Tables~\ref{tab7-2} and~\ref{tab7-3}). 

Duo to unreliable independence tests using a small-sized data samples, a MB learning algorithm using the AND rule may remove the true MB members from its output when performing the symmetry check, and thus this may degrade the performance of  the algorithm on recall. For example, in Table~\ref{tab7-2}, using 500 samples on the two small-sized networks,  PCMB and IPCMB get much lower recall than MMMB, HITON-MB, and semi-HITON-MB.
Thus it is an interesting problem of studying under what conditions we use the AND rule or the OR rule or combining both to make MB learning more accurate.

\item Score-based methods have the similar performance as constraint-based methods in terms of F1.
From Tables~\ref{tab7-4} and~\ref{tab7-5}, S$^{2}$TMB and SLL achieve similar performance as the best constraint-based MB learning methods in terms of  both precision  and recall with the two small-sized networks. These results are consistent with the results of S$^{2}$TMB-PC and SLL-PC in Tables~\ref{tab7-3} and~\ref{tab7-4}. On the large-sized networks, due to expensive computation costs,  S$^{2}$TMB and SLL did not produce results within three days.

\item Simultaneous MB learning methods are the fastest while score-based methods are the slowest.
Since the six simultaneous MB learning methods do not need to perform an exhaustive subset search within the currently selected features, they are faster than the eight divide-and-conquer MB methods. In addition, these six simultaneous MB learning algorithms have the similar performance in terms of efficiency. PCMB and IPCMB are the slowest among the divide-and-conquer MB methods, since they need to perform the symmetry check. Moreover, the Recongize-PC algorithm used by IPCMB is slower than MMPC, HITON-PC, and semi-HITON-PC as shown in Tables~\ref{tab7-2} and~\ref{tab7-3}.
STMB and BAMB do not need to perform the symmetry check. However, they need to  perform an exhaustive subset search for PC learning. The these two algorithms are not significantly faster than MMMB, HITON-MB, and semi-HITON-MB.
S$^{2}$TMB and SLL are slower than all constraint-based MB learning algorithms, since they need to use a BN structure learning algorithm to learn MBs.  S$^{2}$TMB is faster than SLL because S$^{2}$TMB removes the symmetry check.
\end{itemize}

\begin{table}[!htbp]
\centering
\caption{Results of MB learning methods on synthetic data sets (size=500)\label{tab:four}}{
\tiny
\begin{tabular}{cccccccc}
\hline

Data  & Algorithm     & F1        & Precision & Recall    & Time      \\\hline
      & GSMB          & 0.69/0.54 & 0.78/0.59 & 0.67/0.58 & 0/0       \\
      & IAMB          & 0.78/0.67 & 0.86/0.67 & 0.77/0.77 & 0/0       \\
      & Inter-IAMB    & 0.78/0.67 & 0.86/0.67 & 0.77/0.77 & 0/0       \\
      & Fast-IAMB     & 0.74/0.68 & 0.87/0.77 & 0.71/0.70  & 0/0       \\
      & LRH           & 0.80/0.68  & 0.86/0.67 & 0.80/0.82  & 0/0       \\
      & FBED          & 0.79/0.70  & 0.87/0.71 & 0.77/0.77 & 0/0       \\
      & MMMB          & 0.87/0.82 & 0.94/0.84 & 0.83/0.84 & 0/0       \\
      & PCMB          & 0.79/0.79 & 0.93/0.88 & 0.74/0.76 & 0/0       \\
Child & HITON-MB      & 0.85/0.83 & 0.94/0.86 & 0.82/0.85 & 0/0       \\
      & Semi-HITON-MB & 0.85/0.83 & 0.94/0.87 & 0.82/0.85 & 0/0       \\
      & MBOR          & 0.84/0.81 & 0.92/0.83 & 0.81/0.84 & 0/0       \\
      & IPCMB         & 0.80/0.82  & 0.93/0.88 & 0.74/0.79 & 0/0       \\
      & STMB          & 0.81/0.71 & 0.85/0.66 & 0.82/0.86 & 0/0       \\
      & BAMB          & 0.86/0.81 & 0.93/0.82 & 0.83/0.84 & 0/0       \\ 
    & EEMB          & 0.84/0.80  &	0.92/0.81 & 0.82/0.85 & 0/0       \\\cline{2-6}

      & SLL           & 0.84      & 0.95      & 0.79      & 0.85      \\
      & S$^{2}$TMB    & 0.83      & 0.95      & 0.77      & 0.14      \\\hline
      & GSMB          & 0.29/0.19 & 0.35/0.21 & 0.27/0.20  & 0/0       \\
      & IAMB          & 0.76/0.73 & 0.89/0.79 & 0.71/0.74 & 0/0       \\
      & Inter-IAMB    & 0.77/0.72 & 0.89/0.77 & 0.72/0.75 & 0/0       \\
      & Fast-IAMB     & 0.71/0.67 & 0.85/0.77 & 0.65/0.66 & 0/0       \\
      & LRH           & 0.75/0.70  & 0.85/0.71 & 0.73/0.76 & 0/0       \\
      & FBED          & 0.77/0.74 & 0.90/0.80   & 0.71/0.73 & 0/0       \\
      & MMMB          & 0.82/0.82 & 0.90/0.85  & 0.78/0.84 & 0/0       \\
      & PCMB          & 0.74/0.79 & 0.88/0.89 & 0.68/0.76 & 0/0.01    \\
Alarm & HITON-MB      & 0.82/0.81 & 0.90/0.84  & 0.78/0.85 & 0/0       \\
      & Semi-HITON-MB & 0.82/0.83 & 0.90/0.86  & 0.78/0.85 & 0/0       \\
      & MBOR          & 0.85/0.85 & 0.93/0.89 & 0.81/0.85 & 0/0       \\
      & IPCMB         & 0.74/0.80  & 0.87/0.88 & 0.69/0.77 & 0/0.01    \\
      & STMB          & 0.67/0.59 & 0.69/0.54 & 0.76/0.84 & 0/0       \\
      & BAMB          & 0.80/0.80   & 0.90/0.85  & 0.75/0.80  & 0/0       \\
     & EEMB          & 0.80/0.79  & 0.90/0.85  & 0.75/0.87 & 0/0       \\\cline{2-6}

      & SLL           & 0.88      & 0.93      & 0.87      & 0.88      \\
      & S$^{2}$TMB    & 0.87      & 0.92      & 0.86      & 0.29      \\\hline
      & GSMB          & 0.06/0.01 & 0.08/0.02 & 0.05/0.01 & 0/0       \\
      & IAMB          & 0.80/0.80 & 0.98/0.98 & 0.72/0.72 & 0.01/0.01 \\
      & Inter-IAMB    & 0.80/0.80 & 0.98/0.98 & 0.72/0.72 & 0.01/0.01 \\
      & Fast-IAMB     & 0.77/0.76 & 0.80/0.80 & 0.83/0.81 & 0.01/0.01 \\
      & LRH           & 0.69/0.68 & 0.90/0.87 & 0.60/0.60 & 0.01/0.02 \\
      & FBED          & 0.71/0.69 & 0.89/0.87 & 0.64/0.62 & 0/0       \\
      & MMMB          & 0.93/0.76 & 0.88/0.64 & 1/1       & 0.04/0.05 \\
      & PCMB          & 0.99/0.95 & 0.98/0.92 & 1/1       & 0.13/0.17 \\
Pigs  & HITON-MB      & 0.93/0.76 & 0.88/0.64 & 1/1       & 0.07/0.08 \\
      & Semi-HITON-MB & 0.93/0.77 & 0.88/0.65 & 1/1       & 0.08/0.1  \\
      & MBOR          & 0.95/0.85 & 0.92/0.77 & 1/1       & 0.02/0.04 \\
      & IPCMB         & 0.99/0.97 & 0.99/0.95 & 1/1       & 0.60/0.67  \\
      & STMB          & 0.59/0.25 & 0.47/0.15 & 0.97/0.98 & 0.05/0.08 \\
      & BAMB          & 0.96/0.81 & 0.93/0.71 & 1/1       & 0.05/0.06 \\ 
       & EEMB          & 0.96/0.82 & 0.93/0.72 &	1/1	      & 0.04/0.08 \\\cline{2-6}

      & SLL           & -         & -         & -         & -         \\
      & S$^{2}$TMB    & -         & -         & -         & -         \\\hline
      & GSMB          & 0.06/0.01 & 0.08/0.02 & 0.05/0.01 & 0/0       \\
      & IAMB          & 0.66/0.66 & 0.80/0.80 & 0.68/0.68 & 0.01/0.01 \\
      & Inter-IAMB    & 0.66/0.66 & 0.80/0.80 & 0.68/0.68 & 0.01/0.01 \\
      & Fast-IAMB     & 0.66/0.65 & 0.68/0.68 & 0.78/0.78 & 0.01/0.01 \\
      & LRH           & 0.71/0.67 & 0.89/0.82 & 0.67/0.67 & 0.01/0.02 \\
      & FBED          & 0.66/0.66 & 0.80/0.80 & 0.68/0.67 & 0.01/0.01 \\
      & MMMB          & 0.82/0.66 & 0.79/0.56 & 0.91/0.92 & 0.03/0.04 \\
      & PCMB          & 0.94/0.93 & 0.97/0.93 & 0.93/0.95 & 0.08/0.12 \\
Gene  & HITON-MB      & 0.82/0.66 & 0.79/0.56 & 0.91/0.92 & 0.03/0.05 \\
      & Semi-HITON-MB & 0.82/0.67 & 0.79/0.57 & 0.91/0.92 & 0.03/0.05 \\
      & MBOR          & 0.87/0.74 & 0.86/0.66 & 0.91/0.92 & 0.03/0.06 \\
      & IPCMB         & 0.96/0.94 & 0.98/0.94 & 0.94/0.96 & 0.04/0.07 \\
      & STMB          & 0.56/0.18 & 0.44/0.12 & 0.92/0.92 & 0.04/0.06 \\
      & BAMB          & 0.82/0.68 & 0.79/0.59 & 0.90/0.91 & 0.03/0.05 \\
      & EEMB          & 0.82/0.67 & 0.78/0.57 & 0.90/0.92  & 0.03/0.05 \\\cline{2-6}

      & SLL           & -         & -         & -         & -         \\
      & S$^{2}$TMB    & -         & -         & -         & -         \\

\hline
\end{tabular}}
\label{tab7-4}
\end{table}

\begin{table}[!htbp]
\centering
\caption{Results of MB learning methods on synthetic data sets (size=5,000)\label{tab:four}}{
\tiny
\begin{tabular}{cccccccc}
\hline

Data  & Algorithm     & F1        & Precision & Recall    & Time        \\\hline
      & GSMB          & 0.70/0.45 & 0.69/0.42 & 0.75/0.58 & 0/0         \\
      & IAMB          & 0.86/0.72 & 0.83/0.64 & 0.95/0.95 & 0.01/0.01   \\
      & Inter-IAMB    & 0.86/0.72 & 0.83/0.65 & 0.95/0.95 & 0.01/0.01   \\
      & Fast-IAMB     & 0.89/0.74 & 0.89/0.70  & 0.93/0.91 & 0.01/0.01   \\
      & LRH           & 0.85/0.71 & 0.81/0.63 & 0.97/0.96 & 0.02/0.03   \\
      & FBED          & 0.89/0.76 & 0.87/0.69 & 0.94/0.94 & 0.01/0.01   \\
      & MMMB          & 0.98/0.89 & 0.97/0.84 & 1/1       & 0.05/0.05   \\
      & PCMB          & 1/0.98    & 1/0.97    & 1/1       & 0.1/0.11    \\
Child & HITON-MB      & 0.98/0.90  & 0.97/0.85 & 1/1       & 0.06/0.07   \\
      & Semi-HITON-MB & 0.98/0.90  & 0.97/0.85 & 1/1       & 0.07/0.08   \\
      & MBOR          & 0.96/0.88 & 0.96/0.84 & 0.98/0.98 & 0.04/0.08   \\
      & IPCMB         & 1/0.98    & 1/0.97    & 1/1       & 0.09/0.11   \\
      & STMB          & 0.91/0.80 & 0.86/0.71 & 0.99/1    & 0.02/0.03   \\
      & BAMB          & 0.97/0.89 & 0.96/0.83 & 0.99/0.99 & 0.04/0.05   \\ 
      & EEMB          & 0.96/0.88 & 0.95/0.82 & 0.99/0.99 & 0.02/0.02   \\ \cline{2-6}

      & SLL           & 0.96      & 1         & 0.93      & 9.1         \\
      & S$^{2}$TMB    & 0.98      & 1         & 0.96      & 1.3         \\\hline
      & GSMB          & 0.33/0.20  & 0.34/0.20  & 0.35/0.26 & 0.01/0      \\
      & IAMB          & 0.92/0.80  & 0.94/0.75 & 0.91/0.92 & 0.02/0.02   \\
      & Inter-IAMB    & 0.92/0.79 & 0.94/0.72 & 0.92/0.93 & 0.02/0.03   \\
      & Fast-IAMB     & 0.91/0.79 & 0.94/0.75 & 0.91/0.92 & 0.01/0.01   \\
      & LRH           & 0.91/0.78 & 0.93/0.71 & 0.92/0.93 & 0.13/0.05   \\
      & FBED          & 0.93/0.83 & 0.96/0.80  & 0.91/0.92 & 0.01/0.01   \\
      & MMMB          & 0.97/0.93 & 0.98/0.90  & 0.97/0.98 & 0.04/0.04   \\
      & PCMB          & 0.97/0.98 & 1/1       & 0.96/0.96 & 0.08/0.09   \\
Alarm & HITON-MB      & 0.97/0.92 & 0.98/0.89 & 0.97/0.98 & 0.04/0.05   \\
      & Semi-HITON-MB & 0.97/0.93 & 0.98/0.9  & 0.97/0.98 & 0.05/0.06   \\
      & MBOR          & 0.97/0.96 & 0.98/0.95 & 0.97/0.98 & 0.05/0.06   \\
      & IPCMB         & 0.97/0.98 & 1/1       & 0.96/0.97 & 0.09/0.1    \\
      & STMB          & 0.79/0.71 & 0.76/0.63 & 0.95/0.97 & 0.04/0.04   \\
      & BAMB          & 0.96/0.92 & 0.98/0.91 & 0.94/0.96 & 0.03/0.04   \\ 
      & EEMB          & 0.96/0.93 & 0.99/0.93 & 0.94/0.95 & 0.02/0.03   \\\cline{2-6}

      & SLL           & 0.96      & 0.97      & 0.96      & 8.15        \\
      & S$^{2}$TMB    & 0.95      & 0.98      & 0.93      & 2.44        \\\hline
      & GSMB          & 0.05/0.02 & 0.04/0.02 & 0.06/0.03 & 0.02/0.01   \\
      & IAMB          & 0.71/0.71 & 0.62/0.62 & 0.96/0.96 & 0.29/0.29   \\
      & Inter-IAMB    & 0.71/0.71 & 0.62/0.62 & 0.96/0.96 & 0.28/0.28   \\
      & Fast-IAMB     & 0.71/0.70  & 0.62/0.61 & 0.95/0.94 & 0.18/0.14   \\
      & LRH           & 0.80/0.76  & 0.75/0.68 & 0.96/0.96 & 0.87/1.22   \\
      & FBED          & 0.78/0.67 & 0.71/0.59 & 0.95/0.89 & 0.12/0.11   \\
      & MMMB          & 0.92/0.77 & 0.87/0.65 & 1/1       & 6.47/6      \\
      & PCMB          & 0.99/0.97 & 0.99/0.95 & 1/1       & 29.15/23.12 \\
Pigs  & HITON-MB      & 0.92/0.76 & 0.87/0.64 & 1/1       & 9.74/12.27  \\
      & Semi-HITON-MB & 0.92/0.77 & 0.87/0.65 & 1/1       & 13.58/13.6  \\
      & MBOR          & 0.97/0.91 & 0.94/0.86 & 1/1       & 0.73/1.19   \\
      & IPCMB         & 1/0.98    & 1/0.97    & 1/1       & 11.17/11.37 \\
      & STMB          & 0.38/0.15 & 0.28/0.09 & 1/1       & 2.66/5.2    \\
      & BAMB          & 0.96/0.84 & 0.93/0.74 & 1/1       & 23.35/23.28 \\ 
      & EEMB          & 0.96/0.87 & 0.94/0.79 & 1/1	      & 8.71/8.27   \\\cline{2-6}

      & SLL           & -         & -         & -         & -           \\
      & S$^{2}$TMB    & -         & -         & -         & -           \\\hline
      & GSMB          & 0.03/0.01 & 0.03/0.01 & 0.04/0.02 & 0.02/0.01   \\
      & IAMB          & 0.60/0.60   & 0.53/0.53 & 0.89/0.89 & 0.65/0.64   \\
      & Inter-IAMB    & 0.60/0.60   & 0.53/0.53 & 0.89/0.89 & 0.71/0.64   \\
      & Fast-IAMB     & 0.60/0.59  & 0.53/0.52 & 0.89/0.88 & 0.35/0.28   \\
      & LRH           & 0.66/0.61 & 0.60/0.55  & 0.89/0.88 & 0.98/3.3    \\
      & FBED          & 0.62/0.55 & 0.55/0.48 & 0.88/0.82 & 0.24/0.19   \\
      & MMMB          & 0.83/0.68 & 0.77/0.57 & 0.94/0.94 & 0.58/0.69   \\
      & PCMB          & 0.98/0.96 & 0.98/0.95 & 0.99/0.98 & 1.67/2.13   \\
Gene  & HITON-MB      & 0.83/0.68 & 0.77/0.57 & 0.94/0.94 & 0.76/1.1    \\
      & Semi-HITON-MB & 0.83/0.69 & 0.77/0.58 & 0.94/0.94 & 0.73/1.08   \\
      & MBOR          & 0.89/0.83 & 0.86/0.78 & 0.94/0.94 & 0.85/1.83   \\
      & IPCMB         & 0.99/0.97 & 0.99/0.96 & 0.99/0.99 & 1.92/2.96   \\
      & STMB          & 0.30/0.11  & 0.20/0.07  & 0.99/0.98 & 0.95/1.57   \\
      & BAMB          & 0.82/0.69 & 0.76/0.59 & 0.94/0.94 & 0.71/1.55   \\ 
     & EEMB          & 0.82/0.71 & 0.77/0.61 &	0.94/0.94 &	0.56/0.82   \\\cline{2-6}

      & SLL           & -         & -         & -         & -           \\
      & S$^{2}$TMB    & -         & -         & -         & -           \\
\hline
\end{tabular}}
\label{tab7-5}
\end{table}

\subsection{Experiments on real-world data}\label{sec72}

In the section, we validate the causality-based feature selection algorithms using eight real-world datasets from the UCI Machine Learning Repository and NIPS2003 feature selection challenge datasets as shown in Table~\ref{tab7-10}.
Among these eight datasets, three are of low dimensionality but contains a large number of samples, two are high dimensional datasets with large number of samples, and the other three are also of high dimensionality but small size.


We use the three classifiers, NB, KNN, and SVM to evaluate a selected feature subset for classification. The value of $k$ for the KNN classifier is set to 3 and both SVM and KNN use the linear kernel.
In the following, we evaluate two types of feature subsets selected by the causality-based feature selection algorithms for classification, which are the MB and PC of the class variable, using the following metrics.
\begin{itemize}

\item Prediction accuracy. The number of correctly predicted data samples divided by the total number of data samples in a test data set.

\item Compactness. The size of the feature subset selected by an algorithm.

\item AUC. Area Under the ROC Curve for demonstrating the performance of a classifier for predicting a binary class variable with imbalanced classes. 

\end{itemize}

From Tables~\ref{tab7-11} to ~\ref{tab7-13}, our observations are summarized as follows.
\begin{itemize}

\item The classification performance using the PC set of a class variable is not inferior to that of using the MB of the class variable.  And PC learning is much more efficient than MB learning. The findings are consistent with the results in~\citep{aliferis2010local1}. Thus in terms of feature selection, PC learning algorithms are practical in real-world applications, .

\item When the size of data samples is large, the simultaneous MB learning approach (except for GSMB) is significantly faster than the other MB and PC learning algorithms, and they achieve very competitive prediction accuracy with their rivals. Surprisingly, FBED is the fastest algorithm and its performance is comparable with the others as shown in Tables~\ref{tab7-11} to ~\ref{tab7-12}.

\item Existing score-based MB or PC learning algorithms are still computationally expensive or prohibitive when the size of PC or MB is large, even if the number of features in a data set is small. For example, on the  \emph{spambase} data set,  we observe that when the size of MB is over 30, SLL and $S^2$TMB are very computationally expensive using the exact score-based BN learning algorithms.
Using both synthetic and real-world data sets, existing score-based MB or PC learning algorithms do not show significant advantages over the constrain-based algorithms. The results also explain why the score-based MB learning approach is not the focus in the causality-based feature selection research.

\item With the three class-imbalanced data sets, \emph{infant}, \emph{bankruptcy}, and \emph{dorothea}, all MB learning algorithms achieve almost the same high prediction accuracy in Tables~\ref{tab7-11} to ~\ref{tab7-12}. 
In Table~\ref{tab7-13}, on \emph{infant} and \emph{bankruptcy}, it is the same for all PC learning algorithms.
However, in Tables~\ref{tab7-14} to ~\ref{tab7-15}, we can see that all MB and PC learning  algorithms achieve much low values of AUC. Here we do not report the value of AUC of each algorithm on the remaining five class-balanced date sets, since the MB and PC learning  algorithms obtain almost the same prediction accuracy and AUC.  The results indicate that  the existing causality-base feature selection algorithms are not able to deal with a data set with imbalanced classes well.

\item The compactness and symmetry check. 
In terms of compactness, in Tables~\ref{tab7-11} to~\ref{tab7-12}, STMB selects the most features than the other rivals for all data sets, while the score-based MB algorithms achieve the worst performance among all algorithms under comparison. 
Meanwhile, the symmetry check may not be helpful for selecting a good feature subset for classification in real-world applications. MMMB, HITON-MB, and semi-HITON-MB achieve more stable and better prediction accuracy than the other rivals, although their outputs include more false positives than the other rivals as discussed above. The possible explanation is that when we use causality-based feature selection to deal with  real-world data sets for classification, these real-world data sets may violate the  faithfulness or causal sufficiency assumption. 
And this is the same for MMPC, HITON-PC, and semi-HITON-PC in Tables~\ref{tab7-13}. 

\end{itemize}

\begin{table}[!htbp]
\centering
\scriptsize
\caption{Summary of  real-world datasets\label{tab:three}}{%
\begin{tabular}{lrrr}
\hline
Dataset         &	  Number of features    &	Number of instances  &	  Category ratio  \\
\hline

infant          &     86            &   5,337       & 0.94/0.06       \\
spambase        &	  57            &	4,601     	& 0.61/0.39       \\
bankruptcy      &     147           &   7,063       & 0.89/0.11       \\
madelon         &	  500           &	2,000       & 0.50/0.50       \\
gisette         &     5,000         &   6,000       & 0.50/0.50       \\
arcene          &	  10,000        &	100         & 0.56/0.44       \\
dexter          &	  20,000        &	300         & 0.50/0.50       \\
dorothea        &     100,000       &   800         & 0.90/0.10       \\

\hline
\end{tabular}}
\label{tab7-10}
\end{table}

\begin{table}[!htbp]
\centering
\caption{Comparison of MB methods on real-world data sets in prediction accuracy (1)\label{tab:four}}{
\tiny
\begin{tabular}{cccccccc}
\hline

Data     & Algorithm     & NB        & KNN       & SVM       & Compactness  & Time           \\\hline
         & GSMB          & 0.58/0.55 & 0.51/0.52 & 0.56/0.55 & 7.00/7.00    & 0.01/0.00      \\
         & IAMB          & 0.58/0.58 & 0.62/0.62 & 0.63/0.63 & 7.00/7.00    & 0.21/0.22      \\
         & Inter-IAMB    & 0.58/0.58 & 0.62/0.61 & 0.63/0.63 & 7.00/7.00    & 0.22/0.21      \\
         & Fast-IAMB     & 0.57/0.57 & 0.57/0.56 & 0.61/0.60 & 6.00/6.00    & 0.06/0.06      \\
         & LRH           & 0.60/0.61 & 0.61/0.60 & 0.62/0.63 & 8.00/8.00    & 0.16/0.30      \\
         & FBED          & 0.60/0.59 & 0.61/0.57 & 0.62/0.61 & 7.10/7.90    & 0.06/0.03      \\
         & MMMB          & 0.58/0.59 & 0.55/0.63 & 0.60/0.64 & 5.70/9.40    & 0.14/0.23      \\
madelon  & PCMB          & 0.56/0.58 & 0.50/0.52 & 0.55/0.58 & 1.50/3.50    & 0.24/0.42      \\
         & HITON-MB      & 0.58/0.60 & 0.57/0.60 & 0.61/0.63 & 5.60/8.40    & 0.15/0.24      \\
         & Semi-HITON-MB & 0.58/0.60 & 0.56/0.60 & 0.60/0.63 & 5.50/8.30    & 0.15/0.26      \\
         & MBOR          & 0.60/0.60 & 0.59/0.60 & 0.61/0.61 & 6.40/9.40    & 0.38/0.78      \\
         & IPCMB         & 0.59/0.60 & 0.54/0.54 & 0.59/0.61 & 3.70/6.10    & 0.16/0.35      \\
         & STMB          & 0.60/0.60 & 0.56/0.52 & 0.62/0.60 & 24.00/83.40  & 0.13/0.31      \\
         & BAMB          & 0.60/0.59 & 0.61/0.65 & 0.63/0.64 & 7.20/9.60    & 0.27/0.47      \\
         & EEMB          & 0.60/0.61 & 0.59/0.62 & 0.62/0.62 & 6.60/8.10	& 0.17/0.26      \\\cline{2-7}
         & SLL           & 0.57      & 0.52      & 0.57      & 6.1          & 64.47          \\
         & S$^{2}$TMB    & 0.58      & 0.52      & 0.57      & 5.2          & 61.59          \\\hline     
         & GSMB          & 0.71/0.70 & 0.68/0.71 & 0.66/0.67 & 3.60/3.70    & 0.00/0.00      \\
         & IAMB          & 0.69/0.69 & 0.64/0.64 & 0.66/0.66 & 4.00/4.00    & 0.10/0.11      \\
         & Inter-IAMB    & 0.69/0.69 & 0.64/0.64 & 0.66/0.66 & 4.00/4.00    & 0.10/0.11      \\
         & Fast-IAMB     & 0.70/0.67 & 0.66/0.64 & 0.70/0.68 & 3.00/3.00    & 0.05/0.06      \\
         & LRH           & 0.59/0.61 & 0.61/0.58 & 0.59/0.66 & 3.30/4.00    & 0.13/0.30      \\
         & FBED          & 0.68/0.63 & 0.58/0.60 & 0.66/0.65 & 3.90/4.00    & 0.03/0.03      \\
         & MMMB          & 0.80/0.79 & 0.66/0.74 & 0.79/0.76 & 3.90/6.60    & 1.61/2.33      \\
arcene   & PCMB          & 0.61/0.62 & 0.60/0.60 & 0.61/0.63 & 1.70/2.00    & 2.88/5.01      \\
         & HITON-MB      & 0.74/0.70 & 0.67/0.69 & 0.69/0.73 & 3.70/6.90    & 1.71/2.48      \\
         & Semi-HITON-MB & 0.69/0.71 & 0.62/0.62 & 0.66/0.73 & 3.50/5.80    & 1.56/2.20      \\
         & MBOR          & 0.66/0.71 & 0.63/0.73 & 0.66/0.71 & 3.30/6.80    & 0.12/0.59      \\
         & IPCMB         & 0.61/0.63 & 0.61/0.62 & 0.60/0.62 & 1.50/2.80    & 0.57/1.38      \\
         & STMB          & 0.63/0.61 & 0.71/0.71 & 0.71/0.67 & 56.10/293.10 & 0.64/0.87      \\
         & BAMB          & 0.73/0.70 & 0.63/0.65 & 0.66/0.70 & 3.90/7.30    & 0.53/1.28      \\
         & EEMB          & 0.75/0.72 & 0.67/0.66 & 0.69/0.72 & 3.80/5.90    & 0.85/1.13      \\\cline{2-7}
         & SLL           & -         & -         & -         & -            & -              \\
         & S$^{2}$TMB    & 0.74      & 0.7       & 0.73      & 6.1          & 1648.84        \\\hline
         & GSMB          & 0.70/0.63 & 0.67/0.55 & 0.70/0.63 & 4.80/4.60    & 0.00/0.00      \\
         & IAMB          & 0.81/0.81 & 0.73/0.73 & 0.82/0.82 & 5.00/5.00    & 0.36/0.36      \\
         & Inter-IAMB    & 0.81/0.81 & 0.73/0.73 & 0.82/0.82 & 5.00/5.00    & 0.35/0.35      \\
         & Fast-IAMB     & 0.73/0.73 & 0.72/0.74 & 0.76/0.77 & 4.00/4.00    & 0.12/0.13      \\
         & LRH           & 0.78/0.75 & 0.80/0.77 & 0.79/0.77 & 4.80/5.00    & 0.30/1.02      \\
         & FBED          & 0.81/0.81 & 0.73/0.73 & 0.82/0.82 & 5.00/5.00    & 0.05/0.06      \\
         & MMMB          & 0.86/0.88 & 0.84/0.86 & 0.86/0.87 & 9.20/19.40   & 4.10/7.17      \\
dexter   & PCMB          & 0.81/0.85 & 0.71/0.83 & 0.81/0.84 & 6.60/12.30   & 11.74/32.99    \\
         & HITON-MB      & 0.85/0.87 & 0.81/0.86 & 0.85/0.86 & 9.80/19.50   & 4.44/6.99      \\
         & Semi-HITON-MB & 0.85/0.87 & 0.79/0.85 & 0.84/0.86 & 8.40/18.80   & 3.92/6.87      \\
         & MBOR          & 0.91/0.90 & 0.87/0.89 & 0.88/0.89 & 21.60/31.00  & 4.38/10.83     \\
         & IPCMB         & 0.82/0.86 & 0.71/0.85 & 0.82/0.85 & 5.90/12.40   & 6.19/13.08     \\
         & STMB          & 0.88/0.90 & 0.82/0.75 & 0.90/0.90 & 54.50/244.50 & 3.03/9.61      \\
         & BAMB          & 0.86/0.89 & 0.84/0.88 & 0.86/0.88 & 9.90/17.90   & 2.04/3.01      \\ 
         & EEMB          & 0.85/0.87 & 0.82/0.87 & 0.85/0.86 & 9.20/14.20   & 3.32/4.45      \\\cline{2-7}
         & SLL           & -         & -         & -         & -            & -              \\
         & S$^{2}$TMB    & -         & -         & -         & -            & -              \\\hline
         & GSMB          & 0.93/0.90 & 0.92/0.90 & 0.93/0.90 & 4.30/7.00    & 0.05/0.01      \\
         & IAMB          & 0.93/0.93 & 0.93/0.94 & 0.93/0.94 & 7.00/7.00    & 42.48/44.64    \\
         & Inter-IAMB    & 0.93/0.93 & 0.93/0.93 & 0.93/0.93 & 7.00/7.00    & 40.89/41.32    \\
         & Fast-IAMB     & 0.93/0.93 & 0.92/0.90 & 0.93/0.92 & 6.00/6.00    & 12.64/12.89    \\
         & LRH           & 0.94/0.94 & 0.92/0.92 & 0.93/0.93 & 5.90/6.30    & 45.22/303.98   \\
         & FBED          & 0.93/0.93 & 0.93/0.93 & 0.93/0.93 & 7.00/7.00    & 2.72/5.32      \\
         & MMMB          & 0.93/0.93 & 0.92/0.91 & 0.93/0.93 & 8.60/16.60   & 288.00/555.40  \\
dorothea & PCMB          & 0.45/0.92 & 0.45/0.92 & 0.45/0.92 & 0.60/2.50    & 331.04/1259.19 \\
         & HITON-MB      & 0.93/0.93 & 0.91/0.92 & 0.93/0.93 & 10.20/19.90  & 333.89/638.10  \\
         & Semi-HITON-MB & 0.94/0.93 & 0.92/0.92 & 0.93/0.93 & 8.90/14.60   & 302.03/497.61  \\
         & MBOR          & 0.93/0.93 & 0.92/0.93 & 0.93/0.93 & 10.50/31.80  & 171.26/1870.99 \\
         & IPCMB         & 0.36/0.83 & 0.36/0.82 & 0.36/0.83 & 0.50/1.40    & 75.16/538.33   \\
         & STMB          & -/-       & -/-       & -/-       & -/-          & -/-            \\
         & BAMB          & 0.93/0.93 & 0.90/0.91 & 0.93/0.93 & 10.50/20.20  & 329.52/2941.22 \\ 
         & EEMB          & -/-       & -/-       & -/-       & -/-          & -/-           \\\cline{2-7}
         & SLL           & -         & -         & -         & -            & -              \\
         & S$^{2}$TMB    & -         & -         & -         & -            & -              \\

\hline
\end{tabular}}
\label{tab7-11}
\end{table}

\begin{table}[!htbp]
\centering
\caption{Comparison of  MB methods on real-world data sets in prediction accuracy (2)\label{tab:four}}{
\tiny
\begin{tabular}{cccccccc}
\hline

Data      & Algorithm     & NB        & KNN        & SVM       & Compactness   & Time            \\\hline
          & GSMB          & 0.63/0.58 & 0.60/0.55  & 0.63/0.57 & 3.00/3.00     & 0.07/0.07       \\
          & IAMB          & 0.84/0.84 & 0.91/0.91  & 0.91/0.91 & 4.00/4.00     & 5.94/5.86       \\
          & Inter-IAMB    & 0.84/0.84 & 0.91/0.91  & 0.91/0.91 & 4.00/4.00     & 5.82/5.85       \\
          & Fast-IAMB     & 0.84/0.84 & 0.83/0.83  & 0.84/0.84 & 3.00/3.00     & 1.25/1.25       \\
          & LRH           & 0.86/0.86 & 0.90/0.90  & 0.91/0.91 & 6.00/6.00     & 294.35/460.95   \\
          & FBED          & 0.85/0.84 & 0.89/0.91  & 0.90/0.91 & 4.00/4.00     & 3.52/3.69       \\
          & MMMB          & 0.90/0.89 & 0.97/0.96  & 0.75/0.75 & 255.10/334.40 & 332.94/601.17   \\
gisette   & PCMB          & 0.86/0.87 & 0.93/0.95  & 0.87/0.77 & 46.60/113.90  & 1335.92/3115.83 \\
          & HITON-MB      & 0.89/0.88 & 0.97/0.97  & 0.75/0.75 & 361.30/507.50 & 496.91/1035.98  \\
          & Semi-HITON-MB & 0.90/0.89 & 0.97/0.97  & 0.75/0.73 & 315.80/433.50 & 546.27/1117.32  \\
          & MBOR          & -/-       & -/-        & -/-       & -/-           & -/-             \\
          & IPCMB         & 0.86/0.86 & 0.94/0.96  & 0.83/0.73 & 80.90/192.30  & 2904.15/6057.50 \\
          & STMB          & -/-       & -/-        & -/-       & -/-           & -/-             \\
          & BAMB          & -/-       & -/-        & -/-       & -/-           & -/-             \\ 
          & EEMB          & -/-       & -/-        & -/-       & -/-           & -/-             \\\cline{2-7}
          & SLL           & -         & -          & -         & -             & -               \\
          & S$^{2}$TMB    & -         & -          & -         & -             & -               \\\hline
          & GSMB          & 0.89/0.89 & 0.85/0.88  & 0.89/0.89 & 7.20/6.00     & 0.02/0.01       \\
          & IAMB          & 0.89/0.89 & 0.89/0.89  & 0.90/0.90 & 10.00/10.00   & 0.30/0.30       \\
          & Inter-IAMB    & 0.89/0.89 & 0.89/0.89  & 0.90/0.90 & 10.00/10.00   & 0.30/0.31       \\
          & Fast-IAMB     & 0.89/0.89 & 0.86/0.86  & 0.89/0.89 & 9.00/9.00     & 0.08/0.09       \\
          & LRH           & 0.89/0.89 & 0.88/0.88  & 0.89/0.89 & 9.70/9.70     & 4.28/6.26       \\
          & FBED          & 0.89/0.89 & 0.89/0.89  & 0.90/0.90 & 10.00/10.00   & 0.16/0.18       \\
          & MMMB          & 0.84/0.82 & 0.88/0.88  & 0.90/0.89 & 60.70/76.90   & 32.53/79.24     \\
bankrupty & PCMB          & 0.86/0.85 & 0.88/0.89  & 0.89/0.89 & 38.90/54.40   & 99.06/225.14    \\
          & HITON-MB      & 0.84/0.82 & 0.88/0.88  & 0.90/0.89 & 58.30/73.70   & 54.81/132.83    \\
          & Semi-HITON-MB & 0.84/0.83 & 0.88/0.88  & 0.90/0.89 & 55.80/72.30   & 58.64/161.94    \\
          & MBOR          & 0.87/0.85 & 0.88/0.88  & 0.90/0.90 & 32.80/39.90   & 67.79/80.22     \\
          & IPCMB         & 0.86/0.85 & 0.88/0.88  & 0.89/0.89 & 39.70/56.00   & 76.54/243.60    \\
          & STMB          & 0.80/0.80 & 0.88/0.88  & 0.89/0.89 & 89.70/113.80  & 125.16/488.26   \\
          & BAMB          & 0.85/0.84 & 0.89/0.89  & 0.90/0.90 & 42.60/52.80   & 431.61/1523.60  \\ 
          & EEMB          & 0.85/0.85 & 0.89/0.89  & 0.90/0.90 & 37.00/44.90   & 217.67/647.90  \\\cline{2-7}
          & SLL           & 0.88	  & 0.84	   & 0.89	   & 12.22	       & 11737.17        \\
          & S$^{2}$TMB    & 0.89      & 0.83       & 0.89      & 9.1           & 1958.75         \\\hline
          & GSMB          & 0.95/0.95 & 0.95/0.95  & 0.96/0.95 & 4.80/3.40     & 0.02/0.00       \\
          & IAMB          & 0.95/0.95 & 0.95/0.95  & 0.96/0.96 & 5.00/5.00     & 0.06/0.06       \\
          & Inter-IAMB    & 0.95/0.95 & 0.95/0.95  & 0.96/0.96 & 5.00/5.00     & 0.06/0.06       \\
          & Fast-IAMB     & 0.95/0.95 & 0.94/0.94  & 0.96/0.96 & 4.30/4.30     & 0.02/0.02       \\
          & LRH           & 0.95/0.95 & 0.95/0.95  & 0.96/0.96 & 6.00/6.60     & 0.10/0.15       \\
          & FBED          & 0.95/0.95 & 0.95/0.95  & 0.96/0.96 & 5.00/5.00     & 0.02/0.02       \\
          & MMMB          & 0.95/0.95 & 0.95/0.95  & 0.96/0.96 & 6.90/11.30    & 0.12/0.24       \\
infant    & PCMB          & 0.95/0.95 & 0.95/0.95  & 0.96/0.96 & 5.10/7.80     & 0.41/0.93       \\
          & HITON-MB      & 0.95/0.95 & 0.95/0.95  & 0.96/0.96 & 6.40/11.30    & 0.14/0.33       \\
          & Semi-HITON-MB & 0.95/0.95 & 0.95/0.95  & 0.96/0.96 & 6.50/11.30    & 0.18/0.47       \\
          & MBOR          & 0.95/0.95 & 0.95/0.95  & 0.96/0.96 & 7.10/8.30     & 0.30/0.58       \\
          & IPCMB         & 0.95/0.95 & 0.94/0.95  & 0.96/0.96 & 3.50/7.70     & 0.26/0.56       \\
          & STMB          & 0.95/0.94 & 0.95/0.95  & 0.96/0.95 & 14.60/28.00   & 0.13/0.48       \\
          & BAMB          & 0.95/0.95 & 0.95/0.95  & 0.96/0.96 & 5.20/9.20     & 0.17/0.64       \\ 
          & EEMB          & 0.95/0.95 & 0.95/0.95  & 0.96/0.96 & 5.20/8.30	   & 0.09/0.28       \\\cline{2-7}
          & SLL           & 0.96	  & 0.94	   & 0.96	   & 5.17	       & 3087.95         \\
          & S$^{2}$TMB    & 0.95      & 0.94       & 0.95      & 6.5           & 94.77           \\\hline
          & GSMB          & 0.79/0.78 & 0.78/0.78  & 0.80/0.80 & 8.50/8.90     & 0.00/0.00       \\
          & IAMB          & 0.90/0.90 & 0.91/0.91  & 0.91/0.91 & 9.00/9.00     & 0.06/0.06       \\
          & Inter-IAMB    & 0.90/0.90 & 0.91/0.91  & 0.91/0.91 & 9.00/9.00     & 0.06/0.06       \\
          & Fast-IAMB     & 0.88/0.88 & 0.90/0.90  & 0.89/0.89 & 8.00/8.00     & 0.01/0.01       \\
          & LRH           & 0.90/0.90 & 0.91/0.90  & 0.91/0.91 & 8.20/8.90     & 0.88/1.12       \\
          & FBED          & 0.90/0.90 & 0.91/0.91  & 0.91/0.91 & 9.00/9.00     & 0.04/0.05       \\
          & MMMB          & 0.88/0.88 & 0.93/0.93  & 0.93/0.93 & 44.10/49.80   & 12.30/23.11     \\
spambase  & PCMB          & 0.88/0.88 & 0.93/0.93  & 0.93/0.93 & 41.30/47.80   & 58.64/90.65     \\
          & HITON-MB      & 0.88/0.88 & 0.93/0.93  & 0.93/0.93 & 43.10/49.10   & 27.06/52.15     \\
          & Semi-HITON-MB & 0.88/0.88 & 0.93/0.93  & 0.93/0.93 & 42.90/49.00   & 34.29/65.51     \\
          & MBOR          & 0.89/0.89 & 0.92/0.93  & 0.93/0.93 & 37.40/42.90   & 25.74/29.35     \\
          & IPCMB         & 0.88/0.88 & 0.92/0.93  & 0.93/0.93 & 42.00/47.80   & 66.02/111.47    \\
          & STMB          & 0.88/0.89 & 0.93/0.93  & 0.93/0.93 & 50.00/52.90   & 67.55/147.72    \\
          & BAMB          & 0.89/0.89 & 0.92/0.93  & 0.93/0.93 & 35.40/43.30   & 147.33/432.71   \\
          & EEMB          & 0.90/0.90 & 0.92/0.93  & 0.93/0.93 & 31.60/38.00   & 58.51/160.65   \\\cline{2-7}
          & SLL           & -         & -          & -         & -             & -               \\
          & S$^{2}$TMB    & -         & -          & -         & -             & -               \\

\hline
\end{tabular}}
\label{tab7-12}
\end{table}

\begin{table}[!htbp]
\centering
\caption{Comparison of the PC methods on real-world data sets in prediction accuracy\label{tab:four}}{
\tiny
\begin{tabular}{cccccccc}
\hline

Data     & Algorithm     & NB        & KNN       & SVM       & Compactness & Time           \\\hline
         & PC-simple     & 0.60/0.60 & 0.53/0.49 & 0.59/0.60 & 3.60/4.70   & 0.03/0.08      \\
         & MMPC          & 0.58/0.59 & 0.52/0.57 & 0.60/0.60 & 4.70/6.10   & 0.03/0.04      \\
         & HITON-PC      & 0.59/0.59 & 0.52/0.54 & 0.59/0.60 & 4.70/5.90   & 0.03/0.05      \\
madelon  & Semi-HITON-PC & 0.59/0.59 & 0.51/0.54 & 0.59/0.60 & 4.60/5.50   & 0.03/0.05      \\
         & GetPC         & 0.56/0.58 & 0.50/0.51 & 0.56/0.58 & 1.50/2.70   & 0.15/0.23      \\
         & MBtoPC        & 0.60/0.59 & 0.52/0.52 & 0.60/0.60 & 3.40/4.50   & 0.24/0.43      \\\cline{2-7}
         & SLL-PC        & 0.57      & 0.52      & 0.57      & 5.1         & 22.94          \\
         & S$^{2}$TMB-PC & 0.58      & 0.52      & 0.57      & 5.1         & 61.63          \\\hline
         & PC-simple     & 0.61/0.63 & 0.61/0.62 & 0.60/0.61 & 1.50/3.60   & 0.25/0.37      \\
         & MMPC          & 0.78/0.80 & 0.66/0.75 & 0.79/0.77 & 3.80/5.30   & 0.34/0.40      \\
         & HITON-PC      & 0.74/0.72 & 0.67/0.68 & 0.69/0.71 & 3.70/5.60   & 0.34/0.40      \\
arcene   & Semi-HITON-PC & 0.69/0.73 & 0.62/0.61 & 0.66/0.69 & 3.50/4.90   & 0.34/0.40      \\
         & GetPC         & 0.61/0.62 & 0.60/0.60 & 0.61/0.63 & 1.70/2.00   & 1.66/2.42      \\
         & MBtoPC        & 0.66/0.69 & 0.68/0.65 & 0.71/0.69 & 2.70/3.20   & 0.12/0.61      \\\cline{2-7}
         & SLL-PC        & -         & -         & -         & -           & -              \\
         & S$^{2}$TMB-PC & 0.74      & 0.68      & 0.73      & 5.8         & 1633.66        \\\hline
         & PC-simple     & 0.82/0.86 & 0.71/0.84 & 0.82/0.85 & 6.00/10.90  & 1.01/2.06      \\
         & MMPC          & 0.86/0.87 & 0.81/0.87 & 0.86/0.86 & 8.20/13.40  & 0.49/0.63      \\
         & HITON-PC      & 0.85/0.86 & 0.79/0.87 & 0.85/0.86 & 9.00/13.40  & 0.49/0.61      \\
dexter   & Semi-HITON-PC & 0.85/0.86 & 0.78/0.85 & 0.85/0.86 & 7.80/13.00  & 0.49/0.64      \\
         & GetPC         & 0.81/0.85 & 0.72/0.82 & 0.81/0.84 & 6.20/10.60  & 4.16/7.18      \\
         & MBtoPC        & 0.80/0.79 & 0.75/0.75 & 0.82/0.81 & 5.10/5.50   & 4.10/9.95      \\\cline{2-7}
         & SLL-PC        & -         & -         & -         & -           & -              \\
         & S$^{2}$TMB-PC & -         & -         & -         & -           & -              \\\hline
         & PC-simple     & 0.73/0.93 & 0.73/0.92 & 0.73/0.93 & 1.30/2.20   & 36.72/249.97   \\
         & MMPC          & 0.93/0.93 & 0.92/0.91 & 0.93/0.93 & 8.60/15.30  & 28.47/46.96    \\
         & HITON-PC      & 0.93/0.93 & 0.91/0.91 & 0.93/0.93 & 9.90/17.80  & 28.09/42.30    \\
dorothea & Semi-HITON-PC & 0.94/0.94 & 0.92/0.92 & 0.93/0.93 & 8.80/13.90  & 28.04/39.01    \\
         & GetPC         & 0.54/0.92 & 0.54/0.92 & 0.54/0.92 & 0.80/2.20   & 266.76/615.19  \\
         & MBtoPC        & 0.94/0.93 & 0.92/0.92 & 0.93/0.94 & 6.40/7.60   & 165.69/1726.01 \\\cline{2-7}
         & SLL-PC        & -         & -         & -         & -           & -              \\
         & S$^{2}$TMB-PC & -         & -         & -         & -           & -              \\ \hline
          & PC-simple     & 0.89/0.90 & 0.93/0.94 & 0.90/0.87 & 22.30/39.60 & 1154.45/3168.57 \\
          & MMPC          & 0.91/0.91 & 0.96/0.96 & 0.84/0.82 & 53.70/73.90 & 67.47/169.92    \\
          & HITON-PC      & 0.91/0.91 & 0.96/0.97 & 0.83/0.80 & 48.80/67.90 & 121.95/329.02   \\
gisette   & Semi-HITON-PC & 0.91/0.90 & 0.96/0.96 & 0.84/0.82 & 46.50/63.60 & 135.25/372.99   \\
          & GetPC         & 0.87/0.88 & 0.89/0.91 & 0.90/0.88 & 11.50/22.70 & 291.91/640.68   \\
          & MBtoPC        & -/-       & -/-       & -/-       & -/-         & -/-             \\\cline{2-7}
          & SLL-PC        & -         & -         & -         & -           & -               \\
          & S$^{2}$TMB-PC & -         & -         & -         & -           & -               \\\hline
          & PC-simple     & 0.88/0.87 & 0.86/0.87 & 0.89/0.89 & 23.30/31.30 & 29.92/92.30     \\
          & MMPC          & 0.87/0.87 & 0.87/0.87 & 0.89/0.89 & 30.00/37.40 & 12.56/30.69     \\
          & HITON-PC      & 0.87/0.87 & 0.87/0.88 & 0.89/0.89 & 28.50/36.00 & 22.43/56.45     \\
bankrupty & Semi-HITON-PC & 0.88/0.87 & 0.87/0.87 & 0.89/0.89 & 27.00/35.00 & 26.36/72.68     \\
          & GetPC         & 0.89/0.88 & 0.86/0.87 & 0.89/0.89 & 15.40/24.30 & 56.52/163.50    \\
          & MBtoPC        & 0.89/0.89 & 0.88/0.89 & 0.89/0.90 & 8.20/9.90   & 4.05/5.17       \\\cline{2-7}
          & SLL-PC        & 0.89      & 0.84      & 0.89      & 9.3         & 2668.09         \\
          & S$^{2}$TMB-PC & 0.89      & 0.84      & 0.89      & 8.9         & 1984.47         \\\hline
          & PC-simple     & 0.95/0.95 & 0.95/0.95 & 0.96/0.96 & 3.80/6.70   & 0.06/0.14       \\
          & MMPC          & 0.95/0.95 & 0.95/0.95 & 0.96/0.96 & 5.00/7.90   & 0.02/0.06       \\
          & HITON-PC      & 0.95/0.95 & 0.95/0.95 & 0.96/0.96 & 5.00/8.10   & 0.03/0.10       \\
infant    & Semi-HITON-PC & 0.95/0.95 & 0.95/0.95 & 0.96/0.96 & 5.00/7.90   & 0.04/0.15       \\
          & GetPC         & 0.95/0.95 & 0.95/0.95 & 0.96/0.96 & 4.00/6.40   & 0.11/0.25       \\
          & MBtoPC        & 0.95/0.95 & 0.95/0.95 & 0.96/0.96 & 5.00/5.00   & 0.14/0.19       \\\cline{2-7}
          & SLL-PC        & 0.95      & 0.94      & 0.95      & 5.4         & 2653.88         \\
          & S$^{2}$TMB-PC & 0.95      & 0.94      & 0.95      & 5.9         & 94.8            \\\hline
          & PC-simple     & 0.90/0.90 & 0.92/0.92 & 0.93/0.93 & 21.10/26.80 & 19.17/26.59     \\
          & MMPC          & 0.91/0.91 & 0.92/0.92 & 0.93/0.93 & 23.30/28.60 & 2.06/4.08       \\
          & HITON-PC      & 0.90/0.91 & 0.92/0.92 & 0.93/0.93 & 23.30/28.10 & 5.02/9.97       \\
spambase  & Semi-HITON-PC & 0.91/0.91 & 0.92/0.92 & 0.93/0.93 & 23.20/28.10 & 6.18/12.78      \\
          & GetPC         & 0.90/0.91 & 0.92/0.92 & 0.93/0.93 & 21.90/27.10 & 25.86/49.17     \\
          & MBtoPC        & 0.90/0.90 & 0.90/0.91 & 0.91/0.91 & 9.30/9.70   & 0.99/1.02       \\\cline{2-7}
          & SLL-PC        & -         & -         & -         & -           & -               \\
          & S$^{2}$TMB-PC & -         & -         & -         & -           & -               \\
\hline
\end{tabular}}
\label{tab7-13}
\end{table}

\begin{table}[!htbp]
\centering
\caption{Comparison of  MB learning methods on class-imbalanced real-world data in AUC\label{tab:four}}{
\tiny
\begin{tabular}{cccccccc}
\hline

Data     & Algorithm     & NB-AUC    & KNN-AUC   & SVM-AUC   \\\hline

         & GSMB          & 0.73/0.50 & 0.68/0.51 & 0.73/0.50 \\
         & IAMB          & 0.78/0.78 & 0.78/0.78 & 0.82/0.82 \\
         & Inter-IAMB    & 0.78/0.79 & 0.77/0.79 & 0.82/0.82 \\
         & Fast-IAMB     & 0.76/0.76 & 0.74/0.67 & 0.78/0.76 \\
         & LRH           & 0.79/0.78 & 0.75/0.74 & 0.78/0.79 \\
         & FBED          & 0.79/0.79 & 0.78/0.76 & 0.82/0.82 \\
         & MMMB          & 0.75/0.73 & 0.73/0.72 & 0.79/0.77 \\
dorothea & PCMB          & 0.28/0.68 & 0.28/0.68 & 0.28/0.69 \\
         & HITON-MB      & 0.74/0.74 & 0.73/0.73 & 0.77/0.71 \\
         & Semi-HITON-MB & 0.78/0.74 & 0.75/0.72 & 0.79/0.72 \\
         & MBOR          & 0.73/0.78 & 0.72/0.71 & 0.78/0.75 \\
         & IPCMB         & 0.23/0.59 & 0.23/0.57 & 0.23/0.60 \\
         & STMB          & -/-       & -/-       & -/-       \\
         & BAMB          & 0.72/0.73 & 0.68/0.71 & 0.77/0.74 \\
         & EEMB          & -/-       & -/-       & -/-       \\\cline{2-5}
         & SLL           & -         & -         & -         \\
         & S$^{2}$TMB    & -         & -         & -         \\ \hline

          & GSMB          & 0.50/0.50 & 0.51/0.50 & 0.50/0.50 \\
          & IAMB          & 0.55/0.55 & 0.70/0.70 & 0.59/0.59 \\
          & Inter-IAMB    & 0.55/0.55 & 0.70/0.70 & 0.59/0.59 \\
          & Fast-IAMB     & 0.50/0.50 & 0.73/0.70 & 0.50/0.50 \\
          & LRH           & 0.56/0.56 & 0.69/0.69 & 0.52/0.53 \\
          & FBED          & 0.54/0.55 & 0.69/0.70 & 0.59/0.59 \\
          & MMMB          & 0.78/0.78 & 0.66/0.64 & 0.56/0.54 \\
bankrupty & PCMB          & 0.74/0.77 & 0.64/0.66 & 0.52/0.53 \\
          & HITON-MB      & 0.78/0.78 & 0.65/0.65 & 0.56/0.54 \\
          & Semi-HITON-MB & 0.78/0.78 & 0.65/0.65 & 0.56/0.54 \\
          & MBOR          & 0.74/0.76 & 0.65/0.65 & 0.56/0.57 \\
          & IPCMB         & 0.75/0.78 & 0.65/0.66 & 0.54/0.54 \\
          & STMB          & 0.78/0.78 & 0.63/0.63 & 0.53/0.51 \\
          & BAMB          & 0.78/0.79 & 0.68/0.68 & 0.58/0.57 \\
          & EEMB          & 0.77/0.78 & 0.67/0.67 & 0.57/0.57 \\\cline{2-5}
          & SLL           & 0.62      & 0.66      & 0.50      \\
          & S$^{2}$TMB    & 0.51      & 0.73      & 0.5       \\ \hline
          & GSMB          & 0.72/0.65 & 0.59/0.58 & 0.68/0.65 \\
          & IAMB          & 0.74/0.74 & 0.66/0.66 & 0.69/0.69 \\
          & Inter-IAMB    & 0.74/0.74 & 0.66/0.66 & 0.69/0.69 \\
          & Fast-IAMB     & 0.72/0.72 & 0.53/0.53 & 0.68/0.68 \\
          & LRH           & 0.74/0.73 & 0.66/0.65 & 0.69/0.68 \\
          & FBED          & 0.74/0.74 & 0.66/0.66 & 0.69/0.69 \\
          & MMMB          & 0.74/0.74 & 0.67/0.65 & 0.69/0.69 \\
infant    & PCMB          & 0.73/0.73 & 0.64/0.65 & 0.69/0.69 \\
          & HITON-MB      & 0.73/0.74 & 0.66/0.67 & 0.69/0.70 \\
          & Semi-HITON-MB & 0.73/0.74 & 0.66/0.66 & 0.69/0.69 \\
          & MBOR          & 0.74/0.74 & 0.67/0.67 & 0.69/0.69 \\
          & IPCMB         & 0.72/0.73 & 0.55/0.63 & 0.68/0.69 \\
          & STMB          & 0.73/0.74 & 0.66/0.67 & 0.69/0.67 \\
          & BAMB          & 0.74/0.74 & 0.67/0.67 & 0.69/0.69 \\
          & EEMB          & 0.73/0.74 & 0.65/0.66 &	0.69/0.69 \\\cline{2-5}
          & SLL           & 0.71      & 0.50      & 0.71      \\
          & S$^{2}$TMB    & 0.71      & 0.50      & 0.71      \\

\hline
\end{tabular}}
\label{tab7-14}
\end{table}

\begin{table}[!htbp]
\centering
\caption{Comparison of PC learning methods on class-imbalanced real-world data in AUC\label{tab:four}}{
\tiny
\begin{tabular}{cccccccc}
\hline

Data     & Algorithm     & NB-AUC    & KNN-AUC   & SVM-AUC   \\\hline

         & PC-simple     & 0.51/0.71 & 0.51/0.65 & 0.51/0.72 \\
         & MMPC          & 0.75/0.73 & 0.72/0.72 & 0.79/0.76 \\
         & HITON-PC      & 0.74/0.73 & 0.72/0.73 & 0.77/0.73 \\
dorothea & Semi-HITON-PC & 0.78/0.74 & 0.75/0.72 & 0.79/0.74 \\
         & GetPC         & 0.33/0.68 & 0.33/0.67 & 0.33/0.69 \\
         & MBtoPC        & 0.77/0.78 & 0.72/0.73 & 0.81/0.82 \\\cline{2-5}
         & SLL-PC        & -         & -         & -         \\
         & S$^{2}$TMB-PC & -         & -         & -         \\\hline

          & PC-simple     & 0.72/0.74 & 0.64/0.67 & 0.50/0.51 \\
          & MMPC          & 0.76/0.77 & 0.67/0.67 & 0.50/0.51 \\
          & HITON-PC      & 0.74/0.76 & 0.67/0.68 & 0.50/0.51 \\
bankrupty & Semi-HITON-PC & 0.74/0.77 & 0.66/0.67 & 0.50/0.51 \\
          & GetPC         & 0.63/0.70 & 0.65/0.65 & 0.50/0.51 \\
          & MBtoPC        & 0.54/0.55 & 0.67/0.69 & 0.50/0.59 \\\cline{2-5}
          & SLL-PC        & 0.54      & 0.70       & 0.50       \\
          & S$^{2}$TMB-PC & 0.51      & 0.74      & 0.50       \\\hline
          & PC-simple     & 0.73/0.74 & 0.61/0.65 & 0.69/0.69 \\
          & MMPC          & 0.73/0.74 & 0.66/0.66 & 0.69/0.69 \\
          & HITON-PC      & 0.73/0.74 & 0.66/0.66 & 0.69/0.68 \\
infant    & Semi-HITON-PC & 0.73/0.74 & 0.66/0.65 & 0.69/0.68 \\
          & GetPC         & 0.73/0.73 & 0.62/0.62 & 0.69/0.69 \\
          & MBtoPC        & 0.74/0.74 & 0.66/0.66 & 0.69/0.69 \\\cline{2-5}
          & SLL-PC        & 0.71      & 0.50       & 0.71      \\
          & S$^{2}$TMB-PC & 0.71      & 0.50       & 0.71      \\

 \hline
\end{tabular}}
\label{tab7-15}
\end{table}

\section{Conclusion and open problems}\label{sec8}

In this paper, we first reviewed the state-of-the art causality-based feature selection algorithms, then described our developed open-source software package which implements the representative causality-based feature selection algorithms, and finally we evaluated the  representative algorithms using synthetic and real-world data sets.
 The experiments using synthetic data sets have demonstrated that  a backward strategy or the symmetry check significantly improves the precision of causality-based feature selection algorithms (i.e., helps removing more false positives). However, with real-world data sets, due to noise or small-sized data samples, allowing more false positives in the output of causality-based feature selection algorithms may be useful for classification.
 Using real world data sets, selecting a PC of the class variable for classification would be better and is much faster than selecting a MB. The score-based PC or MB learning algorithms do not suffer from the problem of data efficiency compared to the constraint-based MB learning algorithms.

Although a significant number of causality-based  feature selection algorithms have been developed for big data analytics in the past two decades, more efforts are still required in this research direction to tackle the following challenges.
\begin{itemize}

\item Low-quality data.
Missing or noise data are ubiquitous in many real-world application domains, which means that the values for one or more features in a data set are  incorrect  or missing from recorded observations. All existing causality-based feature selection methods assume that all features involved in a dataset do not have missing or noise values. It is challenging to address causality-based feature selection methods with low-quality big data.

\item Streaming data.
Existing causality-based feature selection methods assume that all data instances (or samples) in a data set are given before the  learning starts. In fact, many real-world data sets are available in streams. There is a need to develop online causality-based feature selection algorithms to deal with streaming data.

\item Weak-supervision data.
In real-world applications, a data set may have very few labeled data instances, while abundant unlabeled data instances are available.  However, existing causality-based feature selection algorithms are unable to work well with such data sets. Thus, it is interesting and challenging to exploit unlabeled data instances to facilitate weak-supervision causality-based feature selection methods.

\item Imbalanced class data.
The majority of existing causality-based methods cannot deal with data sets with imbalanced classes, which, however, exist in many real-world applications. It is important to develop new feature selection methods to address this problem.

\item Distribution shift.
 Causality-based feature selection has the potential to deal with data with distribution shifts, i.e., training data and testing data do not have identical distributions, since in this case the set of all direct causes of a class variable is the invariance and minimal set between training data and testing data. However, it is an open problem to distinguish causes from effects of a class variable.

\end{itemize}

\bibliography{cfs.bib}

\end{document}